\documentclass[lettersize,journal]{IEEEtran}
\usepackage{amsmath,amsfonts}
\usepackage{algorithmic}
\usepackage{algorithm}
\usepackage{array}
\usepackage[caption=false,font=normalsize,labelfont=sf,textfont=sf]{subfig}
\usepackage{textcomp}
\usepackage{stfloats}
\usepackage{url}
\usepackage{verbatim}
\usepackage{graphicx}
\usepackage{cite}
\usepackage{bm} % 引入 bm 包以使用 \bm 命令
\usepackage{amsmath} % 用于数学公式
\usepackage{amssymb}  % 提供额外数学符号，例如 \varnothing 和其他集合符号
\usepackage{graphicx} % 用于插入图片
\usepackage{booktabs} % 提供\toprule、\midrule、\bottomrule命令

\newtheorem{assumption}{Assumption}
\newtheorem{remark}{Remark}
\newtheorem{theorem}{Theorem}
\newtheorem{problem}{Problem}
\newtheorem{definition}{Definition}

\usepackage{xcolor}  % 用于更改文中的字体颜色

\usepackage{url} % 用于处理超链接

\begin{document}

\title{Navigating Robot Swarm Through a Virtual Tube with Flow-Adaptive Distribution Control}

\author{Yongwei Zhang, Shuli Lv, Kairong Liu, Quanyi Liang, Quan Quan,~\IEEEmembership{Senior Member,~IEEE}, and Zhikun She
        % <-this % stops a space
\thanks{Yongwei Zhang, Kairong Liu, Quanyi Liang and Zhikun She are with the School of Mathematical Sciences, Beihang University, Beijing 100191, China (e-mail: \{zhangyongwei, krliu, qyliang, zhikun.she\}@buaa.edu.cn).}
% <-this % stops a space
\thanks{Shuli Lv and Quan Quan are with the School of Automation Science and Electrical Engineering, Beihang University, Beijing 100191, China (e-mail: \{lvshuli, qq\_buaa\}@buaa.edu.cn).}
\thanks{Corresponding authors: Quan Quan and Zhikun She.}
\thanks{This work was supported by the National Key Research and Development
Program of China (No. 2022YFA1005103), National Natural Science Foundation of China (NSFC 12371452, NSFC 12401574).}
}



\maketitle

\begin{abstract}
With the rapid development of robot swarm technology and its diverse applications, navigating robot swarms through complex environments has emerged as a critical research direction. 
To ensure safe navigation and avoid potential collisions with obstacles, the concept of virtual tubes has been introduced to define safe and navigable regions. 
However, current control methods in virtual tubes face the congestion issues, particularly in narrow ones with low throughput.
To address these challenges, we first propose a novel control method that combines a modified artificial potential field (APF) for swarm navigation and density feedback control for distribution regulation. 
Then we generate a global velocity field that not only ensures collision-free navigation but also achieves locally input-to-state stability (LISS) for density tracking. 
Finally, numerical simulations and realistic applications validate the effectiveness and advantages of the proposed method in navigating robot swarms through narrow virtual tubes.

\end{abstract}

\begin{IEEEkeywords}
robot swarm, virtual tube, safe navigation, distribution regulation, density feedback control.
\end{IEEEkeywords}

\section{Introduction}
Navigating robot swarms through complex environments is critical for applications like environmental monitoring, drug delivery, and disaster response \cite{chung2018survey, elamvazhuthi2019mean, maffettone2024mixed}. 
In obstacle-dense settings, virtual tubes \cite{mao2024optimal, mao2022making} define safe navigation corridors . Recent work ensures swarm safety within such tubes \cite{gao2022ref1, gao2023ref2}. 
However, narrow tubes with low throughput cause congestion and collision risks. To address this, we propose a control framework integrating swarm distribution regulation with safe navigation for collision-free and efficient traversal.

Safely navigating swarms through narrow virtual tubes requires two key capabilities: adapting to tube geometry and avoiding collisions with boundaries and other robots.  Existing control strategies fall into two categories: bottom-up and top-down \cite{crespi2008top}. Bottom-up methods like formation control \cite{sun2016exponential} and distributed trajectory planning \cite{park2023dlsc} work well for small swarms but become infeasible at larger scales. Control-based approaches including artificial potential fields (APFs) \cite{rodrigues2018APF}, vector fields \cite{rezende2021vector_field}, control barrier functions (CBFs) \cite{borrmann2015CBF}, and flocking \cite{sakai2016flocking} offer low computation/communication costs and fast adaptation to dynamics. However, they struggle to ensure both safety and efficiency in narrow passages \cite{song2024speed}.

Top-down approaches coordinate large swarms via macroscopic density modeling. The continuous-time Markov chain (CTMC) framework partitions environments into sub-regions to achieve desired distributions but ignores individual dynamics \cite{deshmukh2018markov}. For stochastic systems, mean-field methods use Fokker-Planck equations to synthesize velocity fields that guide agents toward target configurations \cite{elamvazhuthi2019bilinear, zheng2021transporting}, while deterministic approaches employ continuity equations from fluid mechanics \cite{eren2017velocity, eren2018density, krishnan2018distributed}. 
Recent advances have introduced more structured and adaptive macroscopic frameworks. Hierarchical control strategies
organize agents into clusters to improve scalability \cite{saravanos2023distributed}, while leader-follower models steer swarms via PDE(partial differential equation)-governed subgroups\cite{maffettone2025leader}.
Optimization-based approaches formulate density control as optimal control problems \cite{chen2023density, sinigaglia2025robust}. 
In parallel, distributed methods stabilize mean-field dynamics using kernel density estimation with dynamic average consensus \cite{zheng2021distributed} or continuification with PI (proportional-integral) consensus \cite{di2025decentralized}.
Despite these advances, most models treat agents as collision-free point particles, neglecting physical size and safety constraints in confined environments.

We propose a control strategy integrating modified artificial potential fields with density feedback techniques. For safe navigation, we employ line-integral Lyapunov functions and barrier functions from \cite{quan2023distributed} to ensure collision-free motion. Swarm distribution is regulated through a continuity equation model \cite{chorin1990fluid} that links individual dynamics to collective behavior. This enables design of a global velocity field that dynamically guides the swarm toward desired spatial distribution which adapts to tube geometry.

While effective for general tubes, the scheme in \cite{quan2023distributed} causes excessive density and safety risks in narrow configurations. Similarly, distribution control methods \cite{deshmukh2018markov, elamvazhuthi2019bilinear, zheng2021transporting, eren2017velocity, eren2018density, krishnan2018distributed} neglect individual safety areas, increasing collision risks. Our framework specifically addresses narrow-tube challenges through distribution-regulated control, enabling safe navigation while guaranteeing density tracking stability. The main contributions of this article are:
\begin{itemize} 
	\item \textbf{Micro-macro dynamics integration:} Both microscopic collision avoidance and macroscopic density evolution are considered. It enables swarm configuration control while preventing inter-robot collisions.
	\item \textbf{Saturated controller design:} By integrating modified APF with density feedback control, a saturated velocity command composed of simple terms is designed, effectively reducing congestion in narrow tubes.
	\item \textbf{Safety and stability guarantees:} A global velocity field is generated, under which safe navigation and LISS density tracking is rigorously proven, validated through simulations and experiments.
\end{itemize}

\section{Preliminaries} \label{preliminary}
% In this section, we first introduce the notation used in this article. Next, we present a general virtual tube model, then introduce the concepts of virtual tube area
% and flow capacity. Later we provide a saturated-velocity robot kinematic model. Finally, we define the spatial density function of robot swarm and develop its evolution model.

\subsection{Notation}
Let $E \subset \mathbb{R}^2$ be a measurable set and $T>0$ be a constant. Set $E_T = E \times (0,T]$.
Denote $L^2(E) = \{ f : \|f\|_{L^2(E)} := \left( \int_E|f(x)|^2dx \right)^{1/2} < \infty \}$, endowed with the norm $\|\cdot\|_{L^2(E)}$. 
Denote $L^\infty(E) = \{ f : \|f\|_{L^\infty(E)} := \mathrm{ess~sup}_{x\in E}|f(x)| < \infty$ \}, endowed with the norm $\|\cdot\|_{L^\infty(E)}$. 
Denote $L^\infty(E_T) = \{ f : \|f\|_{L^\infty(E_T)} := \mathrm{ess~sup}_{(x,t)\in E_T}|f(x,t)| < \infty \}$, endowed with the norm $\|\cdot\|_{L^\infty(E_T)}$. 
$\|\cdot\|$ is the 2-norm of a vector. $\left|\cdot\right|$ is the absolute value of a scalar. The inner product operator is denoted as $\cdot$.
Let $\mathcal{T}$ be a bounded and connected virtual tube in $\mathbb{R}^2$ and $\partial \mathcal{T}$ be its boundary. Set $\mathcal{T}_T = \mathcal{T} \times (0,T]$ and $S(\mathcal{T}_T) = \partial\mathcal{T} \times [0,T]$. 
For a function $f(\mathbf{x},t) : \mathcal{T}_T \to \mathbb{R}$, $\mathbf{x}$ and $t$ are respectively spatial and time variables. Let $x_i$ be the $i$th coordinate of $\mathbf{x}$, denote $\partial_i f =\frac{\partial f}{\partial x_i}$.
$\nabla$ is the gradient operator acting, and $\nabla \cdot$ is the divergence operator, both with respect to the spatial variable $\mathbf{x}$. $(x, y)$ and $ \langle l, r \rangle$ are the coordinate representations in the Cartesian and curvilinear coordinate systems, respectively.

\subsection{General Virtual Tube Model}
In this subsection, we introduce the general virtual tube model in $\mathbb{R}^2$, which consists of two parts: generating curve and cross-section \cite{quan2023distributed}.

\begin{figure}[!t]
	\centering
	\subfloat[]{\includegraphics[width=2.2in]{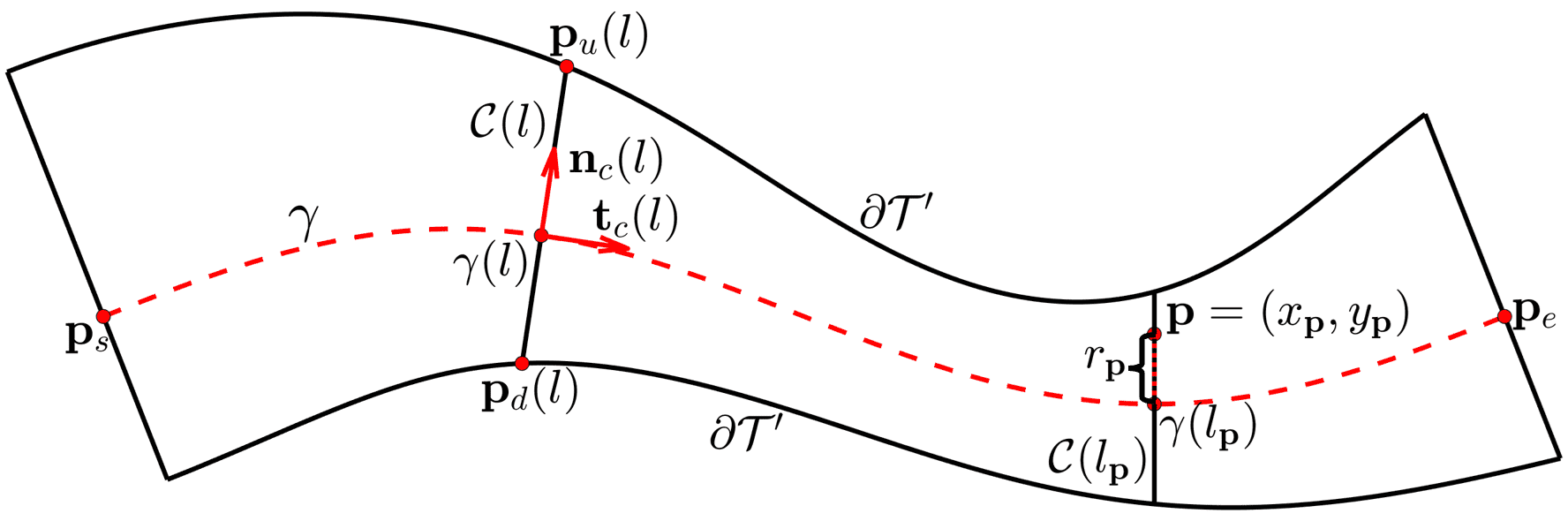}%
			\label{tube}}
	\hfil
	\subfloat[]{\includegraphics[width=1.1in]{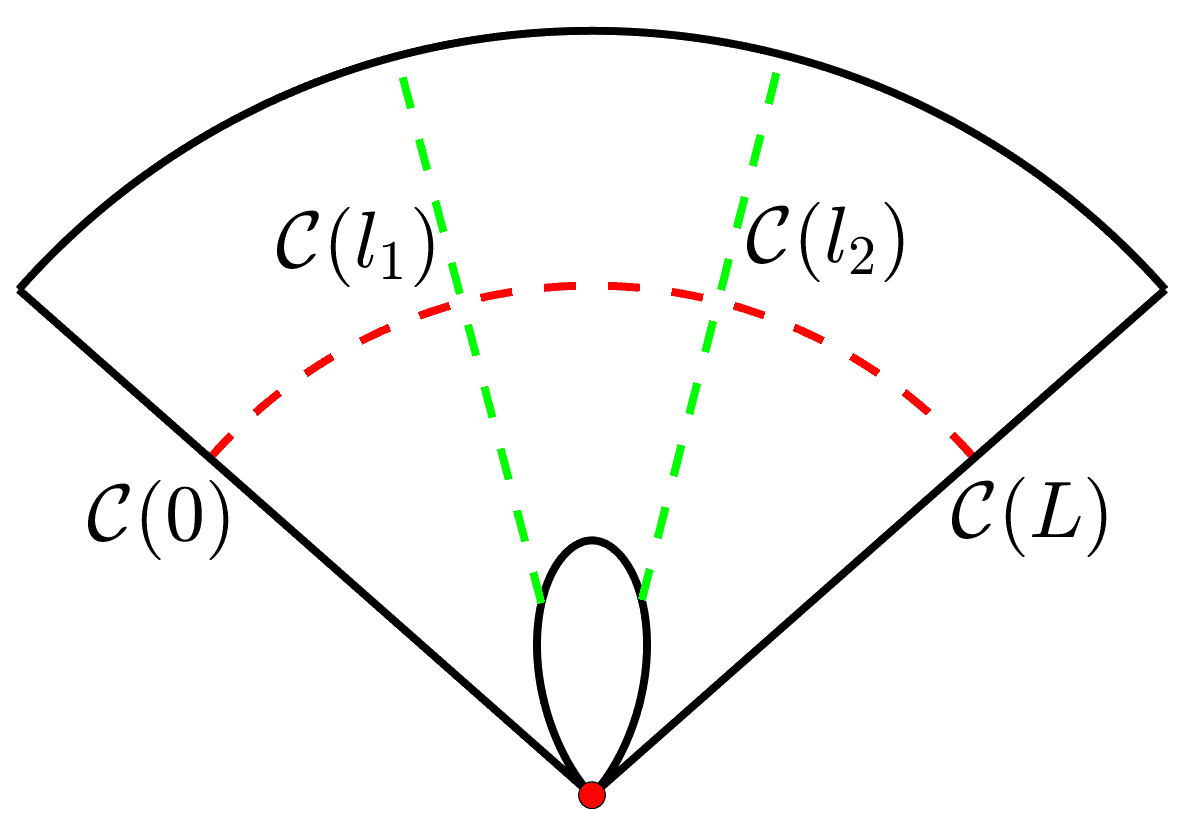}%
			\label{ir_tube}}
	\caption{(a) General regular virtual tube. (b) Irregular virtual tube.}
\end{figure}

1) \textbf{Generating Curve.} Consider a differentiable generating curve $\bm{\gamma}(s)$, where $s \in [0,L]$ denotes the arc length parameter. The curve starts at $\mathbf{p}_s$ and ends at $\mathbf{p}_e$. 
As shown in Fig. \ref{tube}, for a point $\bm{\gamma}(l)$ on the generating curve, $\mathbf{t}_c(l) := \frac{\dot{\bm{\gamma}}(l)}{\left|\dot{\bm{\gamma}}(l) \right|}$ is the unit tangent vector, and $\mathbf{n}_c(l) := \left( -\frac{\dot{\gamma}_2(l)}{\left|\dot{\bm{\gamma}}(l) \right|}, \frac{\dot{\gamma}_1(l)}{\left|\dot{\bm{\gamma}}(l) \right|} \right)$ is the unit normal vector in the counterclockwise direction \cite{ mao2022making, tu2011manifolds}. 

% \begin{figure}[!t] % !t %t
% 	\centering
% 	\includegraphics[width = 2.0in]{figures/tube.png}
% 	\caption{The general virtual tube model.}
% 	\label{tube}
% \end{figure}

2) \textbf{Cross-section.} The cross-section passing $\bm{\gamma}(l)$ is defined as $\mathcal{C}(l) := \{ \mathbf{x} \in \mathbb{R}^2: \mathbf{x} = \mathbf{p}(l) + \lambda(l)\mathbf{n}_c(l), -r_d(l) \le \lambda(l) \le r_u(l), r_d(l), r_u(l) > 0 \}$, where $r_d(l)$ and $r_u(l)$ respectively represent the downward and upward widths of the cross-section. Obviously, two endpoints are $\mathbf{p}_d(l) = \bm{\gamma}(l) - r_d(l)\mathbf{n}_c(l)$ and $\mathbf{p}_u(l) = \bm{\gamma}(l) + r_u(l)\mathbf{n}_c(l)$.

With $\bm{\gamma}(l)$ and $\mathcal{C}(l)$, a general virtual tube is defined as 
\begin{equation}
    \mathcal{T} = \cup_{l \in [0,L]}\mathcal{C}(l)
\end{equation}
Then the boundary of $\mathcal{T}$ is formulated as $\partial \mathcal{T} = \mathcal{C}(0) \cup \mathcal{C}(L) \cup \{ \mathbf{p}_d(l):l \in [0,L] \} \cup \{ \mathbf{p}_u(l):l \in [0,L] \}$,
with the non-terminal boundary $\partial \mathcal{T}' = \partial \mathcal{T} \setminus (\mathcal{C}(0) \cup \mathcal{C}(L))$.
Crucially, the following assumption is required.

\begin{assumption} \label{assumption1}
	For any $l_1, l_2 \in [0,L]$, $l_1 \ne l_2$, $\mathcal{C}(l_1) \cap \mathcal{C}(l_2) = \varnothing$.
\end{assumption}

A virtual tube satisfying Assumption \ref{assumption1} is termed \textit{regular}, defined by non-intersecting cross-sections. Conversely, an \textit{irregular} virtual tube has intersecting cross-sections, as illustrated in Fig. \ref{ir_tube} where terminal sections intersect at a single point. Henceforth $\mathcal{T}$ denotes a regular virtual tube. 

% \begin{figure}[!t]
% 	\centering
% 	\subfloat[]{\includegraphics[width=0.5in]{figures/re_tube.png}}
% 	\hfil
% 	\subfloat[]{\includegraphics[width=0.5in]{figures/irre_tube}}
% 	\caption{(a) Regular virtual tube. (b) Irregular virtual tube.}
% 	\label{r_ir_tube}
% \end{figure}

For any $\mathbf{p} = (x_{\mathbf{p}}, y_{\mathbf{p}}) \in \mathcal{T}$ in Cartesian coordinates, there exists a unique arc length parameter $l_{\mathbf{p}}$ and cross-section $\mathcal{C}(l_{\mathbf{p}})$ that contains $\mathbf{p}$. Define $r_{\mathbf{p}} := \|\mathbf{p} - \bm{\gamma}(l_{\mathbf{p}}) \|$ as the distance to the generating curve. In curvilinear coordinates, $\mathbf{p}$ maps to $\langle l_{\mathbf{p}}, r_{\mathbf{p}} \rangle$ when $(\mathbf{p} - \bm{\gamma}(l_{\mathbf{p}})) \cdot \mathbf{n}_c(l_{\mathbf{p}}) \geq 0$, otherwise to $\langle l_{\mathbf{p}}, -r_{\mathbf{p}} \rangle$.
For $\mathbf{p} \in \mathcal{T}$ with Cartesian $(x_\mathbf{p}, y_\mathbf{p})$ and curvilinear $\langle l_\mathbf{p}, r_\mathbf{p} \rangle$ coordinates, let $\Psi$ be the coordinate transformation and $\pi_1$ the projection operator. By Assumption \ref{assumption1}, $\Psi$ is bijective, yielding the mapping:
\begin{equation} \label{mapping}
	\begin{aligned}
		& \omega :\mathcal{T} \overset{\Psi}{\longrightarrow } \mathcal{T} \overset{\pi_1}{\longrightarrow } \mathbb{R}, \\        
        & \hspace{12pt} (x_\mathbf{p}, y_\mathbf{p}) \longmapsto \langle l_\mathbf{p}, r_\mathbf{p} \rangle \longmapsto l_\mathbf{p},  \\
		&\omega \left((x_\mathbf{p}, y_\mathbf{p})\right)  = \pi_1\left(  \Psi(x_\mathbf{p}, y_\mathbf{p}) \right) = \pi_1(\langle l_\mathbf{p}, r_\mathbf{p} \rangle) = l_\mathbf{p}.
	\end{aligned}
\end{equation}

\subsection{Virtual Tube Area and Flow Capacity}
In this subsection, we define the virtual tube area and flow capacity, essential for macroscopic swarm configuration control. The following assumption is necessary:
\begin{assumption} \label{assumption2}
	For $l \in [0,L]$, assume that $r_d(l)$ and $r_u(l)$ are continuous and differentiable. 
\end{assumption}

Initially, we characterize the virtual tube's spatial capacity by defining its area. Unlike traditional area computations over flat domains, the tube's curved generating curve $\bm{\gamma}(l)$ requires arc-length integration. Specifically, the arc-length coordinate $l$ serves as the integration variable, with the local cross-sectional width $\left(r_d(l) + r_u(l)\right)$ as the integrand. This geometric foundation yields the area definition:
\begin{definition}\label{tube_area}
    The virtual tube area is defined as $S = \int_0^L \left[ r_d(l) + r_u(l) \right] \mathrm{d}l$.
\end{definition}

Subsequently, we define the cross-section radius $r_c(l) := \frac{r_d(l) + r_u(l)}{2}$. The area change rate along the generating curve is given by $\lim_{\Delta l \to 0} \frac{\Delta S}{\Delta l} = 2r_c(l)$, where $\Delta S$ represents the area increment between $l$ and $\left( l+\Delta l \right)$. This result follows from the continuity of $r_c(l)$ and the fundamental theorem of calculus. Since the change rate is proportional to the cross-section radius, we define the flow capacity as:
\begin{definition}\label{flow_capacity}
The flow capacity of cross-section $\mathcal{C}(l)$ is $\sigma(l) := r_c(l)$.
\end{definition}

The flow capacity $\sigma(l)$ directly reflects a cross-section's robot capacity: larger $\sigma(l)$ allows more robots in $\mathcal{C}(l)$. We define $\mathcal{C}(l)$ as \textit{narrow} when $r_s < \sigma(l) \leq 2r_s$, where $r_s$ is the robot \textit{safety radius} \cite{quan2023distributed}. Such narrow cross-sections accommodate at most one robot. A virtual tube containing any narrow cross-section is termed a narrow virtual tube.

\subsection{Single Robot Kinematic Model}
In this subsection, we present the kinematic model for a homogeneous swarm of $N$ robots. Each robot's motion is modeled as
\begin{equation} \label{model}
    \frac{\mathrm{d} \mathbf{p}_i}{\mathrm{d}t}
	 = \mathbf{v}_i, \quad i=1,\cdots,N
\end{equation}
where $\mathbf{p}_i \in \mathbb{R}^2$ is the position, and $\mathbf{v}_i := \mathbf{v}(\mathbf{p}_i,t)$ is the velocity governed by a global vector field $\mathbf{v}$. The velocity saturates due to energy constraints:
\begin{equation}
	\mathbf{v}_i = \mathrm{sat}\left( \mathbf{u}_{i} ,v_{\max} \right) \overset{\mathrm{def}}{=} \kappa_m( \mathbf{u}_{i} ) \mathbf{u}_{i},
\end{equation}
where $\mathbf{u}_{i}$ is the unsaturated velocity.
Both $\mathrm{sat} \left(\mathbf{u}_{i}, v_{\max}\right)$ and $\kappa_m( \mathbf{u}_{i} )$ are defined by comparing $\|\mathbf{u}_{i}\|$ with $v_{\max}$. Specifically, when $\|\mathbf{u}_{i}\|\leq v_{\max}$, $\mathrm{sat} \left(\mathbf{u}_{i}, v_{\max}\right) = \mathbf{u}_{i}$ and $\kappa_m(\mathbf{u}_{i})=1$; when $\|\mathbf{u}_{i}\| > v_{\max}$, $\mathrm{sat} \left(\mathbf{u}_{i}, v_{\max}\right) = v_{\max} \frac{\mathbf{u}_{i}}{\|\mathbf{u}_{i}\|}$ and $\kappa_m(\mathbf{u}_{i})=\frac{v_{\max}}{\|\mathbf{u}_{i}\|}$.
Obviously, $0 < \kappa_m \left(\mathbf{u}_{i}\right) \le 1$, henceforth it will be denoted as $\kappa_m$ for brevity.

% \begin{equation*}
% 	\mathrm{sat} \left(\mathbf{u}_{i}, v_{\max}\right) := \begin{cases}
% 		\hspace{15pt} \mathbf{u}_{i} \hspace{5pt}, & \|\mathbf{u}_{i}\|\leq v_{\max} \\
% 		v_{\max} \frac{\mathbf{u}_{i}}{\|\mathbf{u}_{i}\|}, &\|\mathbf{u}_{i}\|>v_{\max}
% 	\end{cases},
% \end{equation*}
% \begin{equation*}
% 	\kappa_m \left(\mathbf{u}_{i}\right) := \begin{cases}
% 		\hspace{7pt} 1\hspace{7pt}, & \|\mathbf{u}_{i}\|\leq v_{\max} \\
% 		\frac{v_{\max}}{\|\mathbf{u}_{i}\|}, &\|\mathbf{u}_{i}\|>v_{\max}
% 	\end{cases}.
% \end{equation*}

\subsection{Robot Swarm Evolution Model}
In this subsection, we model the swarm's spatial density evolution within $\mathcal{T}$ using the continuity equation \cite{chorin1990fluid}, derived from deterministic robot dynamics.
\begin{definition}[\cite{eren2017velocity, eren2018density, krishnan2018distributed}] \label{density_def}
    For a large swarm of robots ($N \to \infty$), let $\rho: \mathbb{R}^2 \times \mathbb{R}_{\ge 0} \to \mathbb{R}_{\ge 0}$ be the spatial density function supported on $\mathcal{T}$ for $t \in [0,T]$, satisfying $\rho(\mathbf{p},t) \geq 0$ and $\int_\mathcal{T} \rho(\mathbf{p},t) \mathrm{d}A = 1$.
\end{definition}

\begin{remark}
    Following \cite{eren2017velocity,eren2018density}, $\rho(\mathbf{p},t)$ defines a probability density function (PDF) over $\mathcal{T}_T$. Specifically, $\rho(\mathbf{p},t) \mathrm{d}A$ gives the probability of finding a robot within an infinitesimal neighborhood of $\mathbf{p}$ at time $t$, where the neighborhood size approaches zero.
\end{remark}

Assume that $\rho$ is sufficiently smooth and robots follow the deterministic kinematics (\ref{model}), the macroscopic  density evolution follows the continuity equation \cite{eren2017velocity, eren2018density, krishnan2018distributed}:
\begin{subequations} \label{evolution_model}
    \begin{align}
        \frac{\partial \rho}{\partial t} &= -\nabla \cdot (\rho \mathbf{v}), &&(\mathbf{p}, t) \in \mathcal{T}_T \label{evolution_model_a} \\
        \rho &= \rho_0, &&(\mathbf{p}, t) \in \mathcal{T} \times \{0\} \label{evolution_model_b} \\
        \mathbf{n} \cdot (\rho \mathbf{v}) &= 0, &&(\mathbf{p}, t) \in \partial\mathcal{T} \times [0, T] \label{evolution_model_c}
    \end{align}
\end{subequations}

In (\ref{evolution_model_a}), $\mathbf{v}$ is the designed global velocity field, consistent with the individual robot velocities $\mathbf{v}_i(t) = \mathbf{v}(\mathbf{p}_i, t)$ in (\ref{model}). Equation (\ref{evolution_model_b}) specifies the initial density $\rho_0$, while (\ref{evolution_model_c}) imposes a no-flux boundary condition using the outward unit normal $\mathbf{n}$ on $\partial\mathcal{T}$. This condition prevents robots from crossing the tube boundary and conserves the total robot number.

% \int_{\mathcal{B}_\epsilon(\mathbf{p})} \rho(\mathbf{x}, t) \mathrm{d}A  
%  (Theorem 5.4.2 in \cite{Chung2000probability})
For a finite swarm of $N$ robots at positions $\left \{ \mathbf{p}_i(t)\right \}_{i=1}^N$, these positions represent $N$ independent samples from the density $\rho(\mathbf{p},t)$. By the law of large numbers \cite{Chung2000probability}, the probability of a robot residing in an $\epsilon$-neighborhood $\mathcal{B}_\epsilon(\mathbf{p})$ satisfies $\text{Prob}(\mathbf{p}_i \in \mathcal{B}_\epsilon(\mathbf{p}))= 
\mathbb{E}_{\mathbf{x} \sim \rho} [I_{\mathcal{B}_\epsilon(\mathbf{p})}(\mathbf{x})]= \lim_{N \to \infty} \left( \frac{n_{\mathcal{B}_\epsilon(\mathbf{p})}}{N} \right)$, where $I_{\mathcal{B}_\epsilon(\mathbf{p})}(\mathbf{x})$ is the indicator function, equals to 1 if $\mathbf{x} \in \mathcal{B}_\epsilon(\mathbf{p})$ and 0 others; $n_{\mathcal{B}_\epsilon(\mathbf{p})}:=\sum_{i=1}^N I_{\mathcal{B}_\epsilon(\mathbf{p})}(\mathbf{p}_i)$ is the robot number in $\mathcal{B}_\epsilon(\mathbf{p})$. For small $\epsilon$, this probability also approximates $\rho(\mathbf{p}, t) A_{\mathcal{B}_\epsilon(\mathbf{p})}$, with $ A_{\mathcal{B}_\epsilon(\mathbf{p})} $ being the neighborhood area. Equating both approximations yields the density estimator \cite{eren2017velocity, eren2018density}:
\begin{equation}\label{spatial_density}
    \hat{\rho}(\mathbf{p}, t) = \frac{n_{\mathcal{B}_\epsilon(\mathbf{p})}}{N \cdot  A_{\mathcal{B}_\epsilon(\mathbf{p})}} \xrightarrow{N \to \infty, \,\epsilon \to 0} \rho(\mathbf{p}, t)
\end{equation}
Thus, $\rho(\mathbf{p}, t)$ describes both the limiting swarm distribution ($N \to \infty$) and the sampling distribution for finite $N$. The estimator $\hat{\rho}$ becomes accurate for large $N$.

\section{Problem Formulation} 
% In this section, we present the control problem addressed in this article. We start with a high-level definition of control objectives, followed by a mathematical formulation of desired spatial density function.

\subsection{Problem Definition}
Spatial distribution is critical for swarm navigation in virtual tubes. In narrow tubes with limited flow capacity, uncontrolled distribution causes congestion, reduces traversal efficiency, and increases collision risk \cite{song2024speed}. Our objective is therefore to design a control strategy enabling both safe traversal and effective distribution regulation.

Define the $i$th robot's safety area as $ \mathcal{S}_i(t) := \{\mathbf{x} : \|\mathbf{x}-\mathbf{p}_i(t)\| \leq r_s\}$, and let $\rho_d(\mathbf{p},t)$ be the desired density function.

\begin{problem} \label{problem}
	Design the saturated velocity command $\mathbf{v}_i$ in (\ref{model}) to meet the following three requirements:
	\begin{itemize}
		\item Navigate the robot swarm through the regular (possibly narrow) virtual tube. Specifically, for each $\mathbf{p}_i$, $i=1,\cdots,N$, there exists a time $t_i>0$, such that $l_{\mathbf{p}_i(t_i)} = L$.
		
		\item Ensure collision avoidance among robots and with the non-cross-sectional boundary of the virtual tube. Namely, $\mathcal{S}_i(t) \cap \mathcal{S}_j(t) = \varnothing$ and $\mathcal{S}_i(t) \cap \partial\mathcal{T}' = \varnothing$ for $t>0$, $i,j=1,\cdots,N$, $i\ne j$. 
		
		\item The global vector field $\mathbf{v}$ (corresponding to $\mathbf{v}_i$) should drive the solution of (\ref{evolution_model}) towards $\rho_d$, such that the swarm's actual spatial density approaches the desired one.
	\end{itemize}
\end{problem}

\subsection{Desired Spatial Density Function}
\begin{figure}[!t]
	\centering
	\subfloat[]{\includegraphics[width=1.9in]{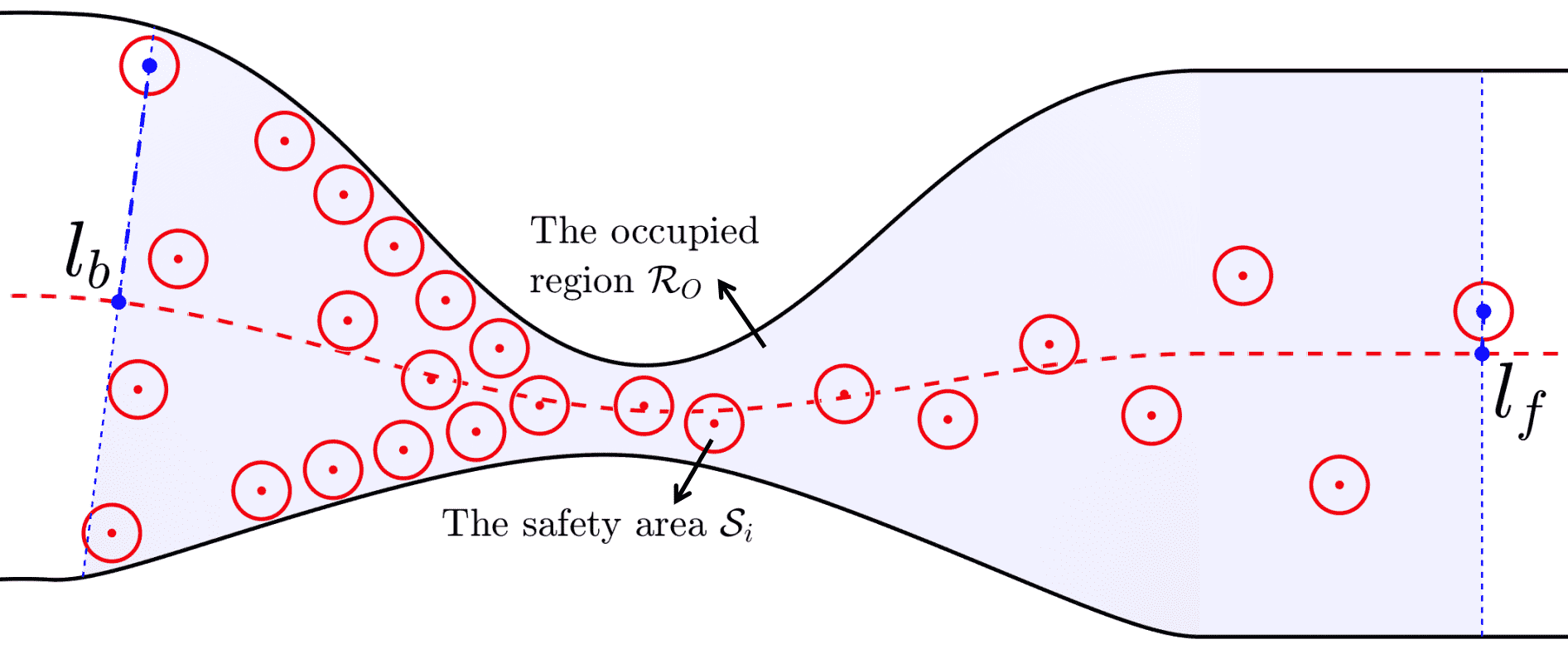}%
			\label{R_O}}
	\hfil
	\subfloat[]{\includegraphics[width=1.5in]{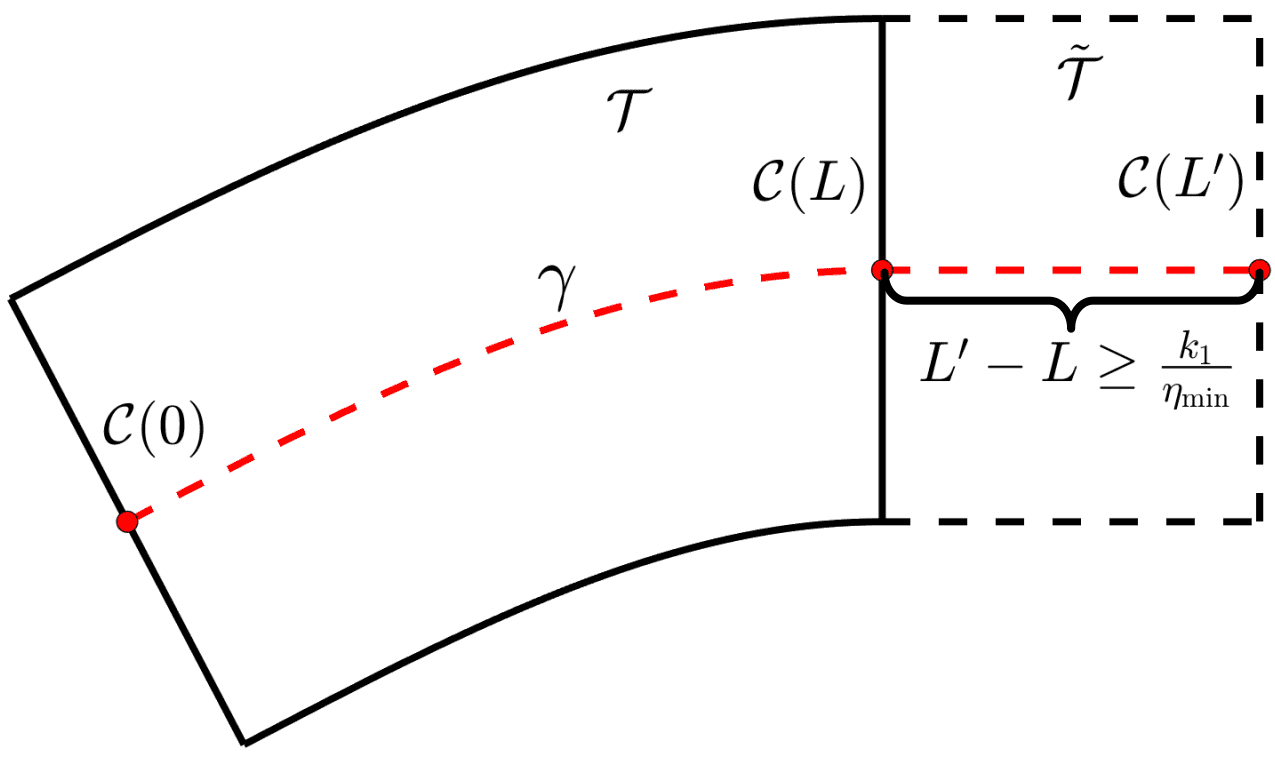}%
			\label{extend}}
	\caption{(a) Schematic diagram for the occupied region $\mathcal{R}_O$. (b) The extended virtual tube $\tilde{\mathcal{T}}$ to modify $\mathbf{u}_{1,i}$.}
\end{figure}

To formulate $\rho_d$ mathematically, we first introduce the key concept of the robot swarm's occupied region. As shown in Fig. \ref{R_O}, the occupied region at time $t$ is defined as $\mathcal{R}_O(t) := \cup_{l \in [l_b(t),l_f(t)]} \mathcal{C}(l)$,
where $l_b(t) = \min_{i=1,\cdots,N} \{l_{\mathbf{p}_i(t)}\}$ and $l_f(t) = \max_{i=1,\cdots,N} \{l_{\mathbf{p}_i(t)}\}$, with $l_{\mathbf{p}_i(t)} = \omega(\mathbf{p}_i(t))$ from the mapping (\ref{mapping}). For notational convenience, the time variable $t$ is omitted henceforth.

% \begin{figure}[!t] 
% 	\centering
% 	\includegraphics[width =3in]{figures/R_O.png}
% 	\caption{Schematic diagram for the occupied region of the robot swarm.}
% 	\label{R_O}
% \end{figure}

% \begin{figure}[!t] 
% 	\centering
% 	\includegraphics[width =2.5in]{figures/extend.png}
% 	\caption{The extend virtual tube $\tilde{\mathcal{T}}$ to modify $\mathbf{u}_{1,i}$}
% 	\label{extend}
% \end{figure}

Next, we design the desired spatial density function $\rho_d$ with the following constraints.
Firstly,
$\rho_d(\mathbf{p},t) \ge 0$ for all $\mathbf{p} \in \mathcal{T}$, and $\int_{\mathcal{T}} \rho_d(\mathbf{p},t) \mathrm{d} A = 1 $, consistent with Definition \ref{density_def};
Secondly,
$\rho_d$ is constant across the same cross-section, i.e., $\rho_d(\mathbf{p}, t) = \rho_d(\mathbf{p}', t)$ if $l_{\mathbf{p}} = l_{\mathbf{p}'}$;
Thirdly,
$\rho_d(\mathbf{p},t) \ne 0$ only within the occupied region $\mathcal{R}_O$, zero otherwise;
Lastly,
for $\mathbf{p} \in \mathcal{R}_O$, $\rho_d(\mathbf{p}, t) = \lambda \sigma(l_{\mathbf{p}}) (\lambda > 0)$, since cross-sections with larger flow capacity accommodate more robots.
In the curvilinear coordinate system, the integral constraint gives $\int_{l_b}^{l_f} \int_{-r_d(l)}^{r_u(l)}\lambda \sigma(l) \mathrm{d}r\mathrm{d}l = 1$.
Solving for $\lambda$ yields $\lambda = \frac{1}{\int_{l_b}^{l_f}2r_c(l)^2\mathrm{d}l}$, thus $\rho_d$ is designed as
\begin{equation}
	\rho_d(\mathbf{p},t) = 
	\begin{cases}
		\frac{r_c(l_{\mathbf{p}})}{\int_{l_b}^{l_f}2r_c(l)^2\mathrm{d}l}, \quad \mathbf{p} \in \mathcal{R}_O,
		\\
		\hspace{18pt} 0, \hspace{18pt} \mathbf{p} \in \mathcal{T} \setminus \mathcal{R}_O.
	\end{cases}
\end{equation}

\section{Controller Design and Guarantee of Safety and Stability}
% In this section, we first employ a modified artificial potential field for safe swarm navigation in the virtual tube. Next, we use density feedback to design a distribution regulation term for configuration control. We then integrate these terms to design a novel saturated controller and derive corresponding global velocity field. Finally, we demonstrate that this field enables collision-free traversal and desired density tracking, solving Problem \ref{problem}.

\subsection{Safe Navigation Design}
In this subsection, we design safe navigation control terms, using three components from our prior work \cite{quan2023distributed}: \textit{Line Approaching}, \textit{Robot Avoidance}, and \textit{Virtual Tube Keeping}, all derived from Lyapunov-like functions.

Firstly, to drive the swarm toward the terminal cross-section $\mathcal{C}(L)$, the Line Approaching term is designed as
\begin{equation}\label{u1}
	\mathbf{u}_{1,i} = \mathrm{sat}  \left((L-l_{\mathbf{p}_i}) \eta(\mathbf{p}_i) \mathbf{t}_c(l_{\mathbf{p}_i}),k_1\right),
\end{equation}
where $k_1$ is the maximum approaching speed, and the associated line integral Lyapunov function \cite{quan2023distributed} is $V_{\mathrm{l},i} = \int_{\mathcal{V}_{\mathbf{p}_i}} \mathrm{sat} \left( -(L-l_\mathbf{x}) \eta(\mathbf{x}), k_1 \right) \mathrm{d} l_\mathbf{x}$.
Here, $\mathcal{V}_{\mathbf{p}_i} $ is the curve from $\mathbf{p}_i$ to $\mathcal{C}(L)$ along $\mathbf{t}_c(\cdot)$, and $\eta(\mathbf{x})$ is the scaling factor with $0 < \eta_{\min} \le \eta(\mathbf{x}) \le \eta_{\max}$.

Secondly, to prevent inter-robot collisions and constrain robots within the virtual tube, the Robot Avoidance and Virtual Tube Keeping terms are respectively:
\begin{equation}
    \begin{aligned}
        \mathbf{u}_{2,i} &= -k_2 \sum_{j\in\mathcal{N}_{\mathrm{m},i}}\frac{\partial V_{\mathrm{m},ij}}{\partial \left\|\tilde{\mathbf{p}}_{\mathrm{m},ij}\right\|}\frac{\tilde{\mathbf{p}}_{\mathrm{m},ij}}{\left\|\tilde{\mathbf{p}}_{\mathrm{m},ij}\right\|}, \\
        \mathbf{u}_{3,i} &= -k_3 \frac{\partial V_{\mathrm{t},i}}{\partial \mathbf{p}_i}.
    \end{aligned}
\end{equation}
where $k_2$ and $k_3$ are positive control gains, $\tilde{\mathbf{p}}_{\mathrm{m},ij} := \mathbf{p}_i - \mathbf{p}_j $ is the relative position between the $i$th and $j$th robots, $ \mathcal{N}_{\mathrm{m},i} := \{ j: \left\| \tilde{\mathbf{p}}_{\mathrm{m},ij} \right\| \le r_s + r_a\}$ denotes the neighbor set of the $i$th robot, with $r_a( > r_s)$ is the \textit{avoidance radius}.
Comprehensive details of $V_{\mathrm{m},ij}$ and $V_{\mathrm{t},i}$ can be found in \cite{quan2023distributed}.

Finally, integrating these three terms gives the \textit{safe navigation} term:
\begin{equation}\label{u123}
    \mathbf{u}_{123,i} = \mathbf{u}_{1,i} + \mathbf{u}_{2,i} + \mathbf{u}_{3,i}.
\end{equation}

\begin{remark}
	The Line Approaching term $\mathbf{u}_{1,i}$ in (\ref{u1}) serves for stability analysis. For practical use, we modify it to $\mathbf{u}_{1,i} = \mathrm{sat} \left((L'-l_{\mathbf{p}_i}) \eta(\mathbf{p}_i) \mathbf{t}_c(l_{\mathbf{p}_i}),k_1\right) = k_1\mathbf{t}_c(l_{\mathbf{p}_i})$,
	where $ L' \ge L + k_1/\eta_{\min} $. 
    This is achieved by extending the virtual tube $\mathcal{T}$ along $\bm{\gamma}$ to $\tilde{\mathcal{T}} = \cup_{l \in [0, L']} \mathcal{C}(l)$, with $\mathcal{C}(L')$ as the modified  terminal cross-section, as shown in Fig. \ref{extend}. 
    In this case, $\mathbf{u}_{1,i}$ within the original $\mathcal{T}$ has a constant magnitude to simplify design and implementation.
    \hfill $\square$
\end{remark}

% \begin{figure}[!t] 
% 	\centering
% 	\includegraphics[width =2.5in]{figures/extend.png}
% 	\caption{The extend virtual tube $\tilde{\mathcal{T}}$ to modify $\mathbf{u}_{1,i}$}
% 	\label{extend}
% \end{figure}

\subsection{Distribution Regulation Design}
In this subsection, we introduce a new distribution regulation term to control the swarm's macroscopic configuration. While the safe navigation term (\ref{u123}) works in general virtual tubes, its lack of distribution regulation causes low traversal efficiency and safety risks in narrow tubes. Thus, regulating the swarm's distribution within the tube is critical.

The density evolution model (\ref{evolution_model}) uses a PDE to describe the swarm's collective dynamics, linking individual and global behaviors. We aim to design the velocity field $\mathbf{v}$ such that $\rho$ converges to $\rho_d$. Define the density tracking error as $\Phi(\mathbf{p},t) = \rho(\mathbf{p},t) - \rho_d(\mathbf{p},t)$, denote $\Phi_0(\mathbf{p}) = \rho_0(\mathbf{p})  - \rho_d(\mathbf{p},0)$. Substitute $\rho = \Phi + \rho_d$ into (\ref{evolution_model}), then $\Phi$ satisfies
\begin{subequations} \label{error_system}
	\begin{align}
		\frac{\partial}{\partial t}(\Phi+\rho_d) &= -\nabla\cdot\left( (\Phi + \rho_d) \mathbf{v} \right), \hspace{5pt}(\mathbf{p},t) \in \mathcal{T}_T; \\ 
		\Phi &= \Phi_0, \hspace{46pt}(\mathbf{p},t) \in \mathcal{T} \times \{0\}; \\
		\mathbf{n} \cdot \left( (\Phi + \rho_d) \mathbf{v} \right)&=0, \hspace{59pt} (\mathbf{p},t) \in S\left(\mathcal{T}_T\right).
	\end{align}
\end{subequations}

Inspired by the diffusion-based stabilization approach in \cite{eren2017velocity, lv2024mean}, we design $\mathbf{v}$ such that the density error $\Phi$ in (\ref{error_system}) satisfies the diffusion equation
\begin{equation} \label{diffusion_equation}
\frac{\partial \Phi(\mathbf{p},t)}{\partial t} = \nabla \cdot (\alpha(\mathbf{p},t) \nabla \Phi(\mathbf{p},t)), \quad (\mathbf{p},t) \in \mathcal{T}_T
\end{equation}
with Neumann boundary condition $\mathbf{n} \cdot \nabla \Phi = 0$ on $S(\mathcal{T}_T) $, where $\alpha(\mathbf{p},t)>0$ is a space-time dependent diffusion coefficient. 
Given the mild condition $\alpha_{\min}(t) := \inf_{\mathbf{p} \in \mathcal{T}} \alpha(\mathbf{p},t) > 0$, it follows that $\|\Phi\|_{L^2(\mathcal{T})} \to 0$, obtained by applying the divergence theorem on $ \int_\mathcal{T} \Phi^2 \mathrm{d}A (= \|\Phi\|_{L^2(\mathcal{T})}^2) $ and leveraging mass conservation, i.e., $\int_\mathcal{T} \Phi(\mathbf{p},t) \mathrm{d} A \equiv 0$ for any $t \geq 0$.

Assume that $\rho(\mathbf{p},t) >0$ and is accurately measurable, the velocity field is designed as
\begin{equation} \label{velocity_field1}
\mathbf{v}(\mathbf{p},t) = -\frac{\alpha(\mathbf{p},t)\left( \nabla \Phi(\mathbf{p},t) + \mathbf{w}_m(\mathbf{p},t) \right)}{\rho(\mathbf{p},t)},
\end{equation}
where $\mathbf{w}_m$ the correction term satisfying $\|\mathbf{w}_m\|_{L^{\infty}(\mathcal{T}_T)} < \infty$ and the constraints:
\begin{equation}\label{constraints_wm}
	\begin{aligned}
		\nabla \cdot \left(\alpha\mathbf{w}_m(\mathbf{p},t) \right) &= \frac{\partial \rho_d(\mathbf{p},t)}{\partial t},  \hspace{15pt}(\mathbf{p},t) \in \mathcal{T}_T; \\
		\mathbf{n} \cdot \mathbf{w}_m(\mathbf{p},t) &= 0, \hspace{33pt}(\mathbf{p},t) \in S\left(\mathcal{T}_T\right).
	\end{aligned}
\end{equation}

Substituting (\ref{velocity_field1}) and (\ref{constraints_wm}) into (\ref{error_system}) yields the diffusion dynamics (\ref{diffusion_equation}) with Neumann boundary condition. Consequently, $\lim_{t\to\infty} \Phi(\mathbf{p},t) = 0$ almost everywhere, meaning $\rho \to \rho_d$ uniformly in $\mathcal{T}$.

However, accurate density estimation requires $N \to \infty$, (see (\ref{spatial_density})), which is impractical. For finite $N$, the density $\rho(\mathbf{p},t)$ estimated via (\ref{spatial_density}) lacks spatial smoothness, leading to high variance in computations. Thus, smooth estimation techniques are needed.
Kernel density estimation (KDE), a nonparametric method \cite{silverman2018density}, estimates $\rho(\mathbf{p},t)$ using $N$ position samples $\{ \mathbf{p}_i(t) \}_{i=1}^N$:
\begin{equation} \label{estimation}
	\hat{\rho}(\mathbf{p},t) = \frac{1}{Nh^2}\sum_{i=1}^N K\left(\frac{\mathbf{p}-\mathbf{p}_i(t)}{h}\right),
\end{equation}
where $h$ is the bandwidth, and $K(\mathbf{x}) = \frac{1}{2\pi}\mathrm{exp}\left( -\frac{1}{2}\mathbf{x}^T\mathbf{x} \right)$ is a smooth Gaussian kernel. For $t \ge 0$, $\mathrm{lim}_{N \to \infty}\| \hat{\rho}(\mathbf{p},t) - \rho(\mathbf{p},t) \|_{L^\infty(\mathcal{T})} = 0$ with probability 1, so estimation accuracy improves with larger $N$.

Using the estimation (\ref{estimation}), the global velocity field (\ref{velocity_field1}) is modified to be
\begin{equation} \label{velocity_field2}
	\mathbf{v}(\mathbf{p},t)= -\frac{\alpha(\mathbf{p},t) \nabla \left(\hat{\rho}(\mathbf{p},t) - \rho_d(\mathbf{p},t) \right)}{\hat{\rho}(\mathbf{p},t)}.
\end{equation}
Note that we set $\mathbf{w}_m \equiv 0$ here to simplify design and implementation. 
Applying (\ref{velocity_field2}) to the $i$th robot gives the \textit{distribution regulation} term:
\begin{equation} \label{u4}
	\mathbf{u}_{4,i}(\mathbf{p}_i,t) = -\frac{\alpha(\mathbf{p}_i,t) \nabla\left( \hat{\rho}(\mathbf{p}_i,t) - \rho_d(\mathbf{p}_i,t) \right) }{\hat{\rho}(\mathbf{p}_i,t)}.
\end{equation}

% The gradient of $\rho_d$ is
% \begin{equation*}
% 	\nabla \rho_d(\mathbf{p},t) = 
% 	\begin{cases}
% 		\frac{\frac{\mathrm{d}}{\mathrm{d}l} r_c(l)|_{l=l_{\mathbf{p}}} \nabla \omega(\mathbf{p})}{\int_{l_b}^{l_f}2r_c(l)^2\mathrm{d}l}, \quad \mathbf{p} \in \mathcal{R}_O,
% 		\\
% 		\hspace{28pt} 0, \hspace{28pt} \mathbf{p} \in \mathcal{T} \setminus \mathcal{R}_O.
% 	\end{cases}
% \end{equation*}

\subsection{Controller Design}
In this subsection, we integrate the safe navigation term (\ref{u123}) and distribution regulation term (\ref{u4}) to design a novel saturated controller. Specifically, we define the unsaturated control input as $\mathbf{u}_{i} := \mathbf{u}_{123,i} + \mathbf{u}_{4,i}$, and the velocity command for (\ref{model}) is
\begin{equation} \label{controller}
	\mathbf{v}_i = \mathrm{sat}\left( \mathbf{u}_{i}, v_{\max} \right)= \kappa_m(\mathbf{u}_{123,i} + \mathbf{u}_{4,i}).
\end{equation}
The $i$th robot's $\mathbf{v}_i$ is in fact derived from the generalized global velocity field:
\begin{equation} \label{velocity_field3}
    \tilde{\mathbf{v}}(\mathbf{p},t) = -\kappa_m
    \frac{\alpha(\mathbf{p},t) \nabla \left( \hat{\rho}(\mathbf{p},t) - \rho_d(\mathbf{p},t) \right)  - \hat{\rho}(\mathbf{p},t)\mathbf{w}_e(\mathbf{p},t)}{\hat{\rho}(\mathbf{p},t)},
\end{equation}
where $\mathbf{w}_e$ is an external vector field satisfying $\mathbf{w}_e(\mathbf{p}_i,t) = \mathbf{u}_{123,i}$. Thus, $\mathbf{v}_i$ can be seen as the result of $\tilde{\mathbf{v}}$ acting on $\mathbf{p}_{i}$.

\begin{remark}
Our proposed control strategy uses a hybrid structure: distributed safe navigation with centralized density regulation. The safe navigation term $\mathbf{u}_{123,i}$ relies on local sensing of neighboring robots and tube boundaries, together with global coordinate information, thus distributed. In contrast, the density regulation term $\mathbf{u}_{4,i}$ requires global position data to compute the KDE $\hat{\rho}(\mathbf{p}_i,t)$, implementable via a centralized monitor.
Recent works achieve distributed density control via local KDE combined with dynamic average consensus \cite{zheng2021distributed} or PI consensus \cite{di2025decentralized}, but require additional assumptions such as connected communication graphs and consensus convergence. Extending our controller to a distributed architecture for safe navigation and stable density regulation remains an important future direction.
\hfill $\square$
\end{remark}

\subsection{Guarantee of safety and stability}
In this subsection, we present the main result of this article. Prior to this, some assumptions are needed.
\begin{assumption} \label{assumption3}
    All robots are initially collision-free and inside the virtual tube: for $i,j = 1,\cdots,N$ with $i \ne j$, $\mathcal{S}_i(0) \cap \mathcal{S}_j(0) = \varnothing$ and $\mathcal{S}_i(0) \cap \partial \mathcal{T} = \varnothing $.
\end{assumption}
\begin{assumption} \label{assumption4}
    Upon reaching the terminal cross-section $\mathcal{C}(L)$, mathematically $l_{\mathbf{p}_i} = L$, the $i$th robot exits the virtual tube immediately and no longer affects other robots’ motion.
\end{assumption}

Next, we present the theorem guaranteeing safe navigation and stable density tracking.
\begin{theorem} \label{theorem}
	For robots modeled by (\ref{model}) under controller (\ref{controller}), assume $\alpha\left( \mathbf{p}_i,t \right)$ in (\ref{u4}) satisfies  
	\begin{equation} \label{condition1}
		\max_{\mathbf{p}_i} \left( \left\| \mathbf{u}_{4,i}(\mathbf{p}_i, t) \right\| - \left\| \mathbf{u}_{123,i}(\mathbf{p}_i, t) \right\| \right) \le 0.
	\end{equation}
    This guarantees safe navigation: $\mathcal{S}_i(t) \cap \partial\mathcal{T}' = \varnothing$ and $\mathcal{S}_i(t) \cap \mathcal{S}_j(t) = \varnothing$ for all robots.
	Moreover, under the global velocity field $\tilde{\mathbf{v}}$ from  (\ref{controller}), the density tracking error $\Phi$ in (\ref{error_system}) is LISS if $\rho_0 > 0$ and
	\begin{equation} \label{condition2}
		\left\|\frac{\nabla\varepsilon}{1+\varepsilon} \right\|_{L^\infty(\mathcal{T})} < \frac{\alpha_{\min}\theta}{C\|\alpha\|_{L^\infty(\mathcal{T})}},
	\end{equation}
	where $\varepsilon(\mathbf{p},t) := \hat{\rho}(\mathbf{p},t)/\rho(\mathbf{p},t) - 1$ is the estimation error, and $C>0$, $\theta\in (0,1)$ are constants. 
\end{theorem}

\textit{Proof:} Please see Appendix \ref{Appendix}.

Theorem \ref{theorem} demonstrates that by integrating the distribution regulation term with the safe navigation term, the proposed controller $\mathbf{v}_i$ enables collision-free swarm navigation through the virtual tube, provided $\mathbf{u}_{4,i}$ satisfies (\ref{condition1}). 
Furthermore, the global vector field $\tilde{\mathbf{v}}$ ensures the density tracking error $\Phi$ is LISS. 
When the estimation error $\varepsilon$ satisfies (\ref{condition2}), $\Phi$ is $L^2$-bounded, implying that the actual density $\rho$ remains close to $\rho_d$ and therefore approximates the desired distribution.
In summary, Theorem \ref{theorem} confirms all requirements of Problem \ref{problem} are met, validating the controller’s effectiveness.

\section{Simulation Results} 
% In this section, we validate the proposed method through simulations and experiments. First, numerical simulations demonstrate safe navigation and stable density tracking in narrow virtual tubes. Next, simulation comparisons highlight the advantages of the distribution regulation term via performance comparisons. Finally, realistic applications verify the method on ground mobile robots using the \textit{Robotarium} platform.
\subsection{Numerical Simulations}
\begin{figure}[!t]
	\centering
	\includegraphics[width = 3.5in]{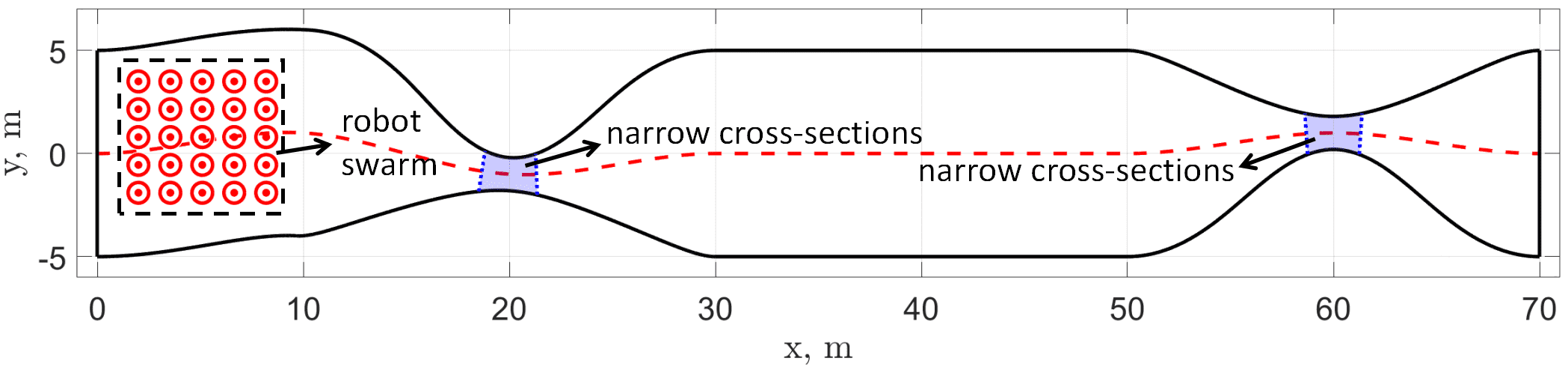}
	\caption{The narrow regular virtual tube and initial distribution of the robots} 
	\label{A0}
\end{figure}

% \begin{table}[!t]
% 	\centering
% 	\caption{Simulation Parameter Settings}
% 	\begin{tabular}{cccccccc}
% 		\toprule % 上线
% 		$k_1$ & $k_2$ & $k_3$ & $r_s$ & $r_a$ & $h$ & $\varepsilon_m/\varepsilon_s/\varepsilon_t$ & $v_{\max}$ \\
% 		\midrule % 中线
% 		2 & 1 & 1 & $0.5m$ & $0.7m$ & 2 & $10^{-6}$ & $3m/s$ \\
% 		\bottomrule % 下线
% 	\end{tabular}
% 	\label{parameter}
% \end{table}
% All parameter settings are provided in Table \ref{parameter}.

In this subsection, we verify the proposed method's effectiveness in a narrow virtual tube environment, as depicted in Figure \ref{A0}.
All robots adopt the kinematic model (\ref{model}). In the figure, the centers of red circles represent robots' current positions, and the radius of these circles indicates the safety radius $r_s$, which is set to $0.5m$.
An initial swarm of $N=25$ robots is arranged in a square formation inside the tube, with no collisions occurring among robots. The narrow cross-sections are highlighted with light blue shading in the figure. The code for reproduction
can be found in our GitHub repository.\footnote{\url{https://github.com/Yongwei-Zhang/Narrow-Virtual-Tube-Navigation}}

\begin{figure}[!t] 
	\centering
	\includegraphics[width = 3.5in]{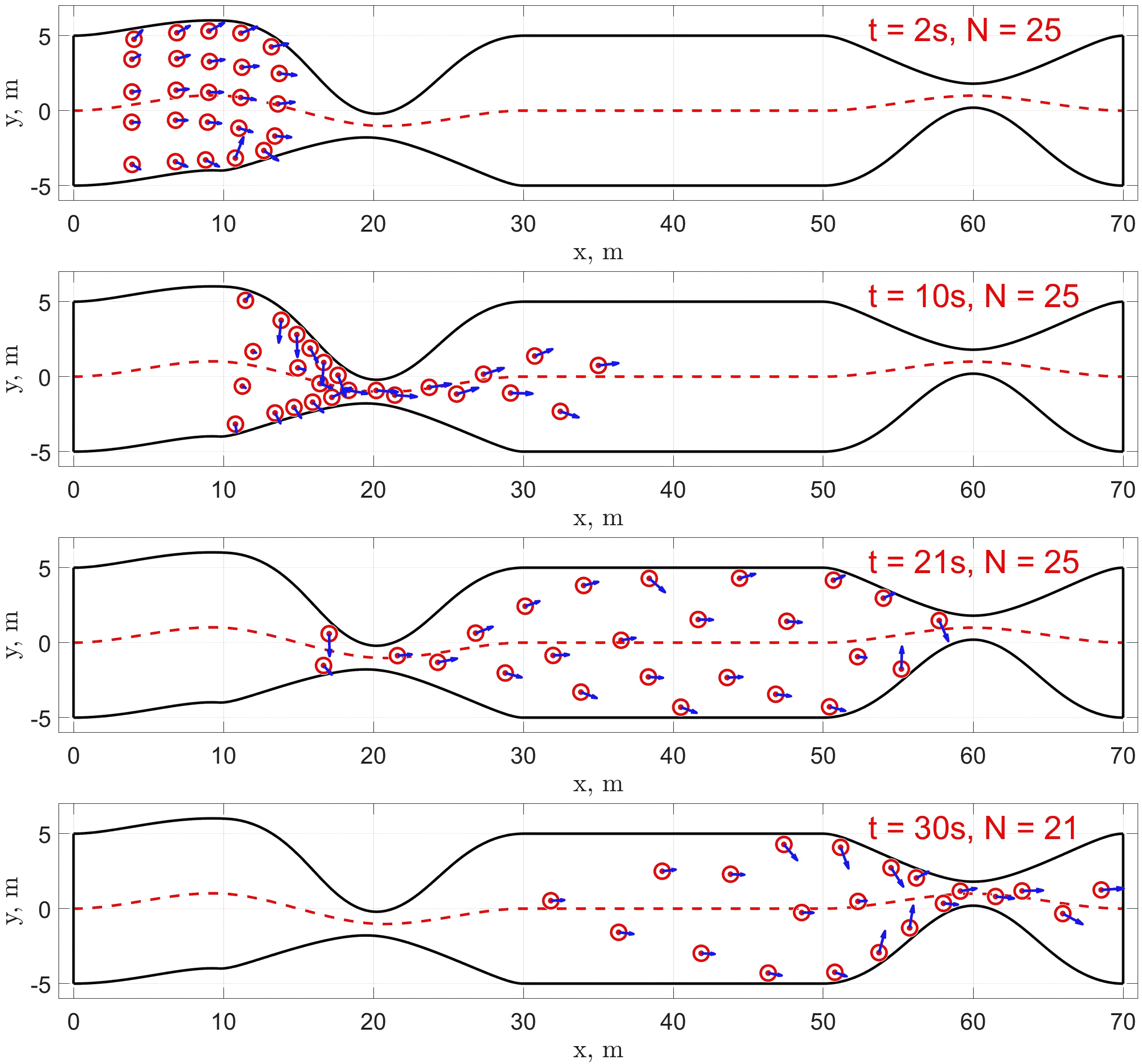}
	\caption{Behavior of robots under the controller $\mathbf{v}_i$}
	\label{A2-30}
\end{figure}

\begin{figure}[!t] 
	\centering
	\includegraphics[width = 3.4in]{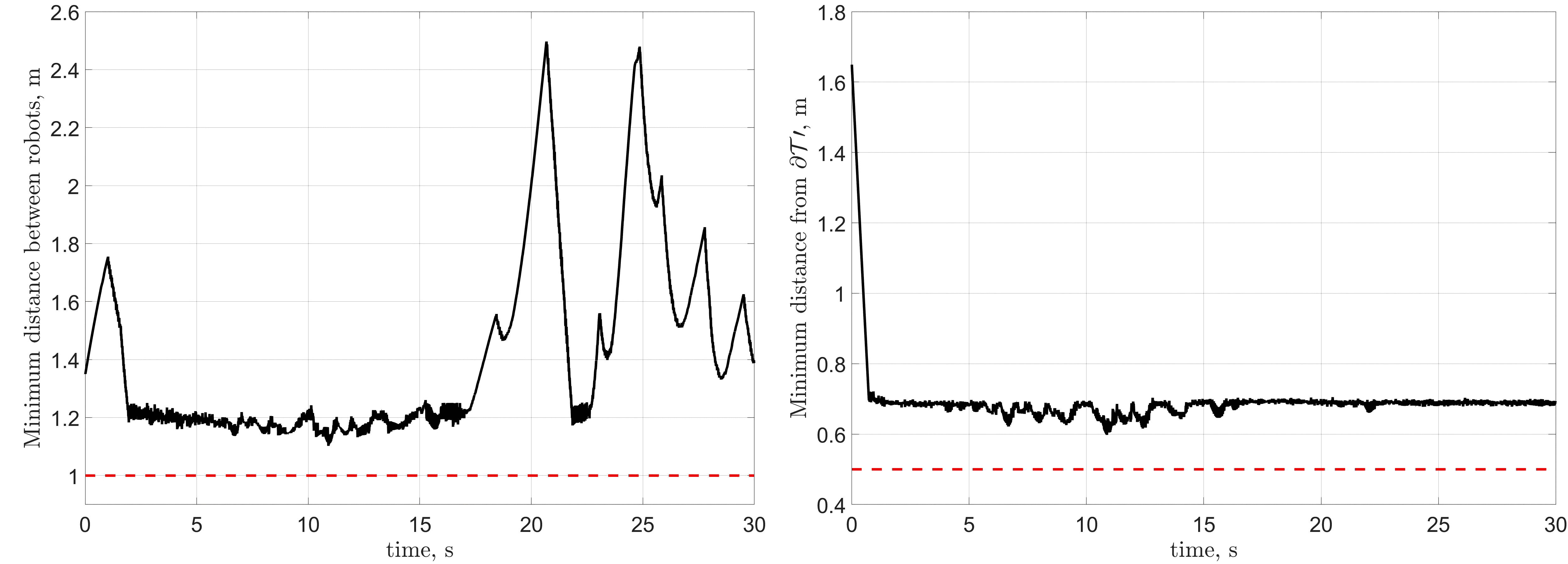}
	\caption{Minimum distance between robots (left) and from $\partial \mathcal{T}'$ (right).}
	\label{robot_tube_distance}
\end{figure}

Screenshots from the 30-second simulation are shown in Fig. \ref{A2-30}, where blue arrows represent the velocity of individual robots. As illustrated in Fig. \ref{robot_tube_distance}, the distance between any two robots remains greater than $2r_s = \,1m$ throughout the simulation, and the minimum distance from robots to $\partial \mathcal{T}'$ stays greater than $r_s = 0.5 \,m$. This confirms that all robots maintain positions within the tube without colliding with each other or the tube boundaries.

\begin{figure}[!t] 
	\centering
	\includegraphics[width = 3.5in]{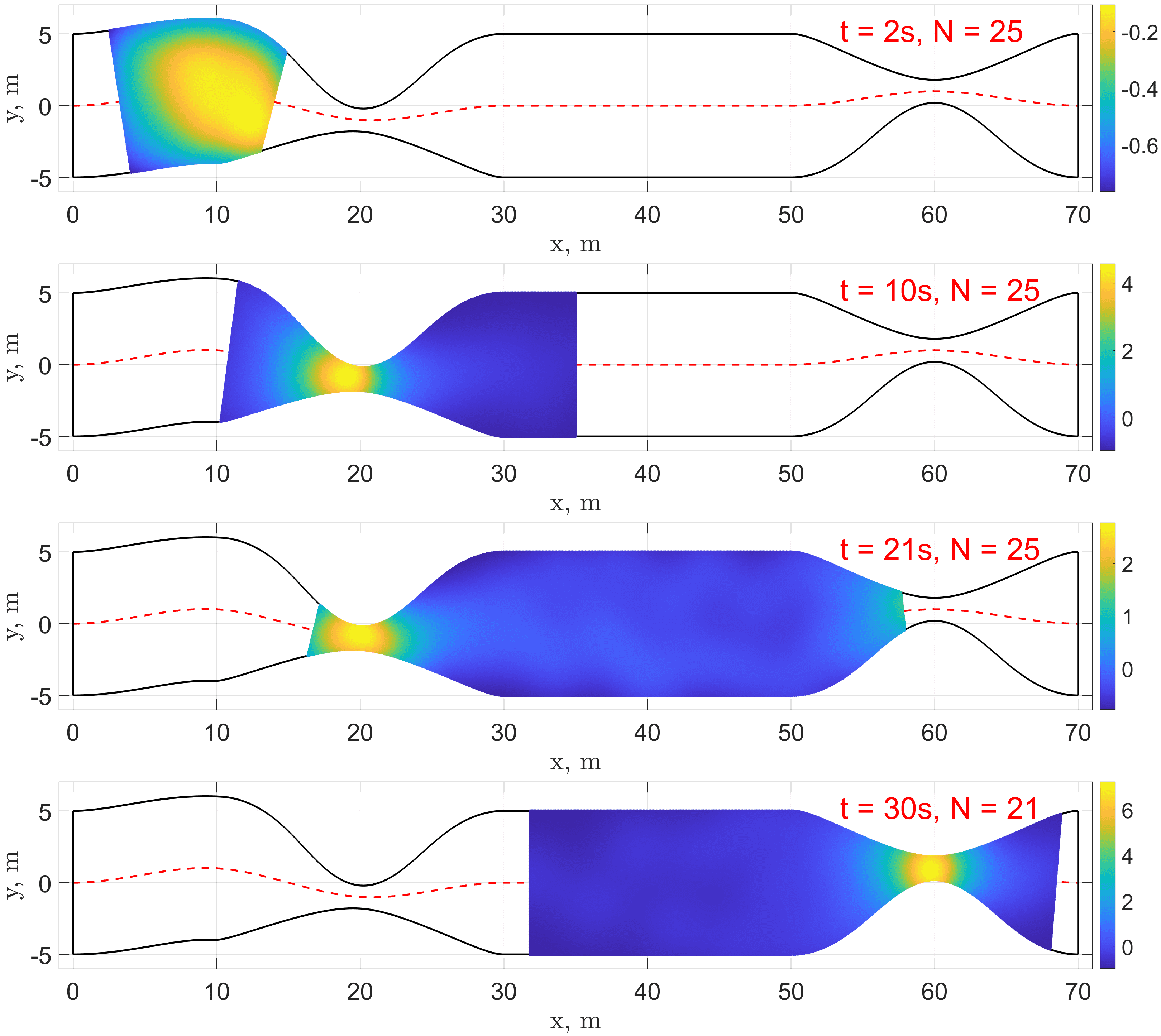}
	\caption{Normalized relative density error between $\hat{\rho}(\mathbf{p},t)$ and $\rho_d(\mathbf{p},t)$}
	\label{E2-30}
\end{figure}

\begin{figure}[!t]
	\centering
	\subfloat[]{\includegraphics[width=1.6in]{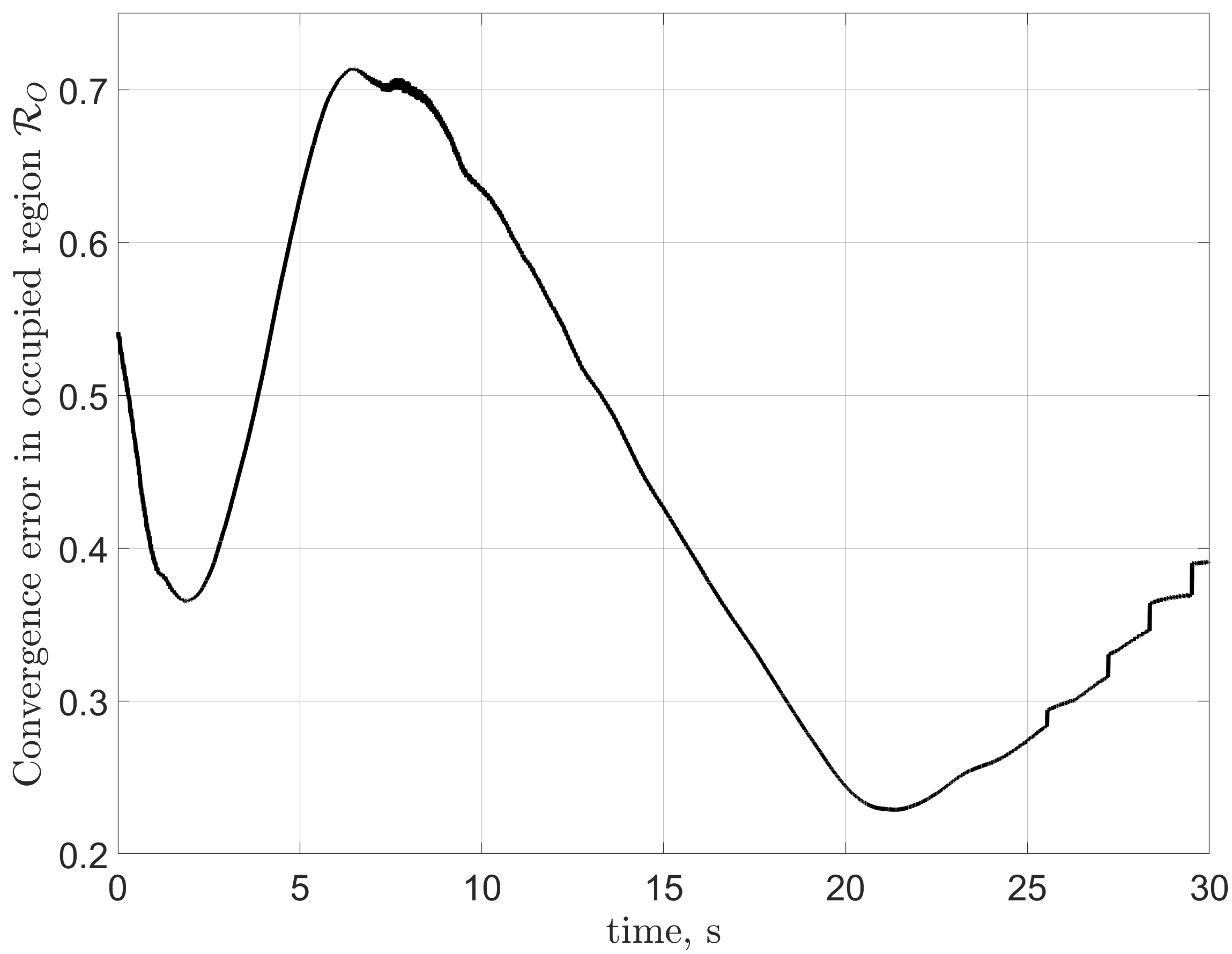}%
		\label{convergence_error}}
	\hfil
	\subfloat[]{\includegraphics[width=1.6in]{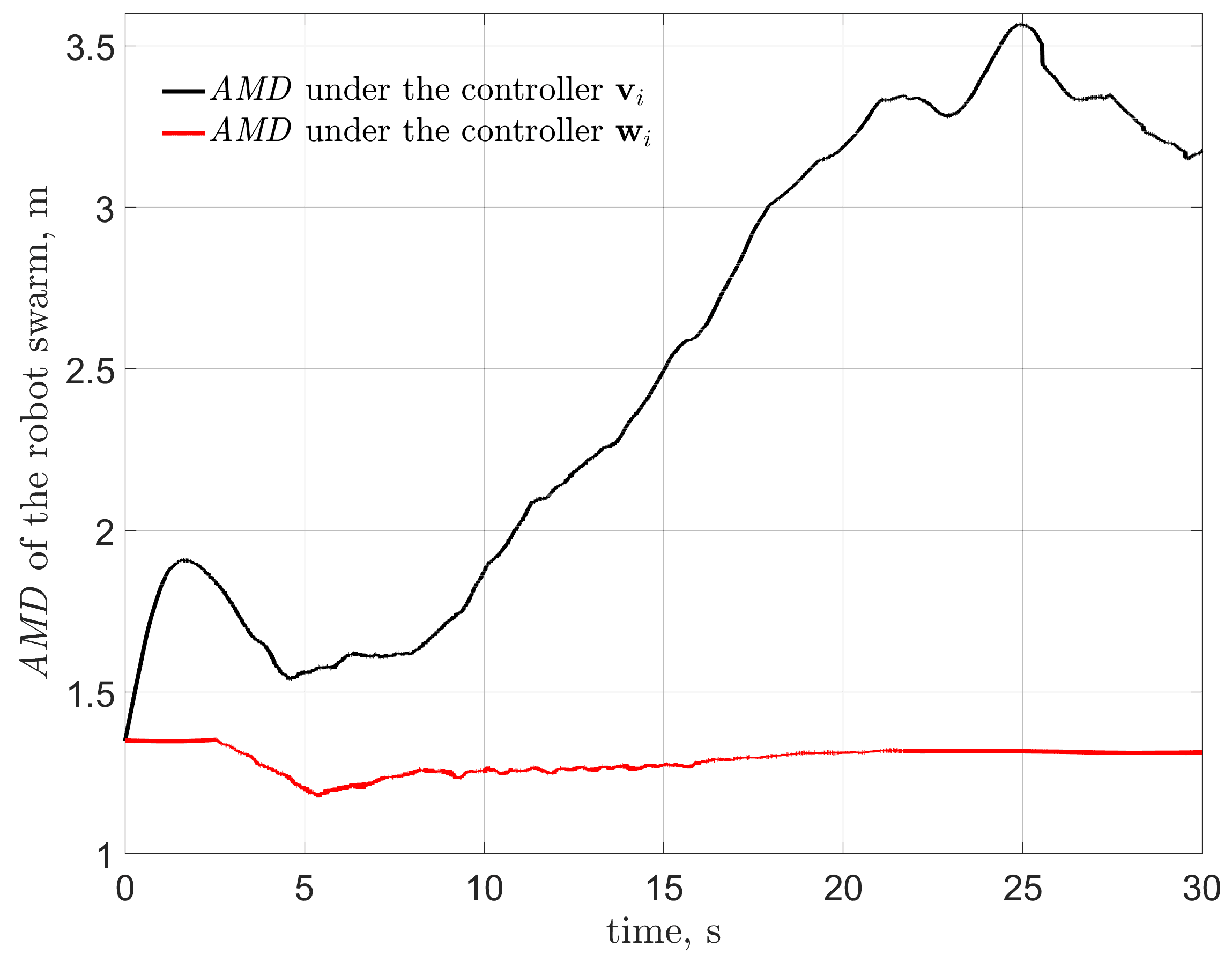}%
		\label{AMD_compare}}
	\caption{(a) Evolution of the convergence error $\|\hat{\rho}-\rho_d\|_{L^2(\mathcal{R}_O)}$. (b) Evolution of AMD under controllers $\mathbf{v}_i$ and $\mathbf{w}_i$.}
\end{figure}

Fig. \ref{E2-30} demonstrates the normalized relative error $\frac{\hat{\rho}(\mathbf{p},t)-\rho_d(\mathbf{p},t)}{\rho_d(\mathbf{p},t)}$ within $\mathcal{R}_O$.
Fig. \ref{convergence_error} presents the convergence error $\|\hat{\rho}-\rho_d\|_{L^2(\mathcal{R}_O)}$, which converges to a small neighborhood of 0 and remains bounded. This result verifies the LISS property of the density tracking errors.

% \begin{figure}[!t]
% 	\centering
% 	\subfloat[]{\includegraphics[width=1.6in]{figures/robot_distance.png}%
% 			% \label{robot_distance}
%             }
% 	\hfil
% 	\subfloat[]{\includegraphics[width=1.6in]{figures/tube_distance.png}%
% 			% \label{tube_distance}
%             }
% 	\caption{(a) Minimum distance between robots. (b) Minimum distance from $\partial \mathcal{T}'$.}
% \end{figure}

\subsection{Simulation Comparisons} \label{simulation_compare}
% \begin{equation}\label{AMD}
% 	AMD = \frac{\sum_{i=1}^N \min_{ j\in \{1,\cdots,N\}, j\ne i }\left\|\tilde{\mathbf{p}}_{\mathrm{m},ij}\right\|}{N}.
% \end{equation}
In this subsection, we use comparative simulations to highlight the advantages of incorporating the distribution regulation term into the controller in \cite{quan2023distributed}. To measure the swarm’s dispersion in the virtual tube, we define the average minimum distance (\textit{AMD}) as
\begin{equation*}
    AMD = \frac{\sum_{i=1}^N \min_{ j\in \{1,\cdots,N\}, j\ne i }\left\|\tilde{\mathbf{p}}_{\mathrm{m},ij}\right\|}{N}
\end{equation*}
% $AMD = \frac{\sum_{i=1}^N \min_{ j\in \{1,\cdots,N\}, j\ne i }\left\|\tilde{\mathbf{p}}_{\mathrm{m},ij}\right\|}{N}$.
% Analogous to highway traffic: a larger \textit{AMD} means greater dispersion, reducing congestion in narrow spaces and improving traversal efficiency and safety.

When the swarm’s distribution is uncontrolled, we use the controller $\mathbf{w}_i := \kappa_m\mathbf{u}_{123,i}$. Figure \ref{AMD_compare} compares AMD values under $\mathbf{v}_i$ and $\mathbf{w}_i$. By examining the simulation results at $t = 30 \,s$, we observe that four robots have exited the tube under $\mathbf{v}_i$, while none have exited under $\mathbf{w}_i$. These results show the distribution regulation term effectively increases AMD, reducing collision risk and improving traversal efficiency.

\subsection{Realistic applications to ground mobile robots}
\begin{figure}[!t] 
	\centering
	\includegraphics[width = 3.5in]{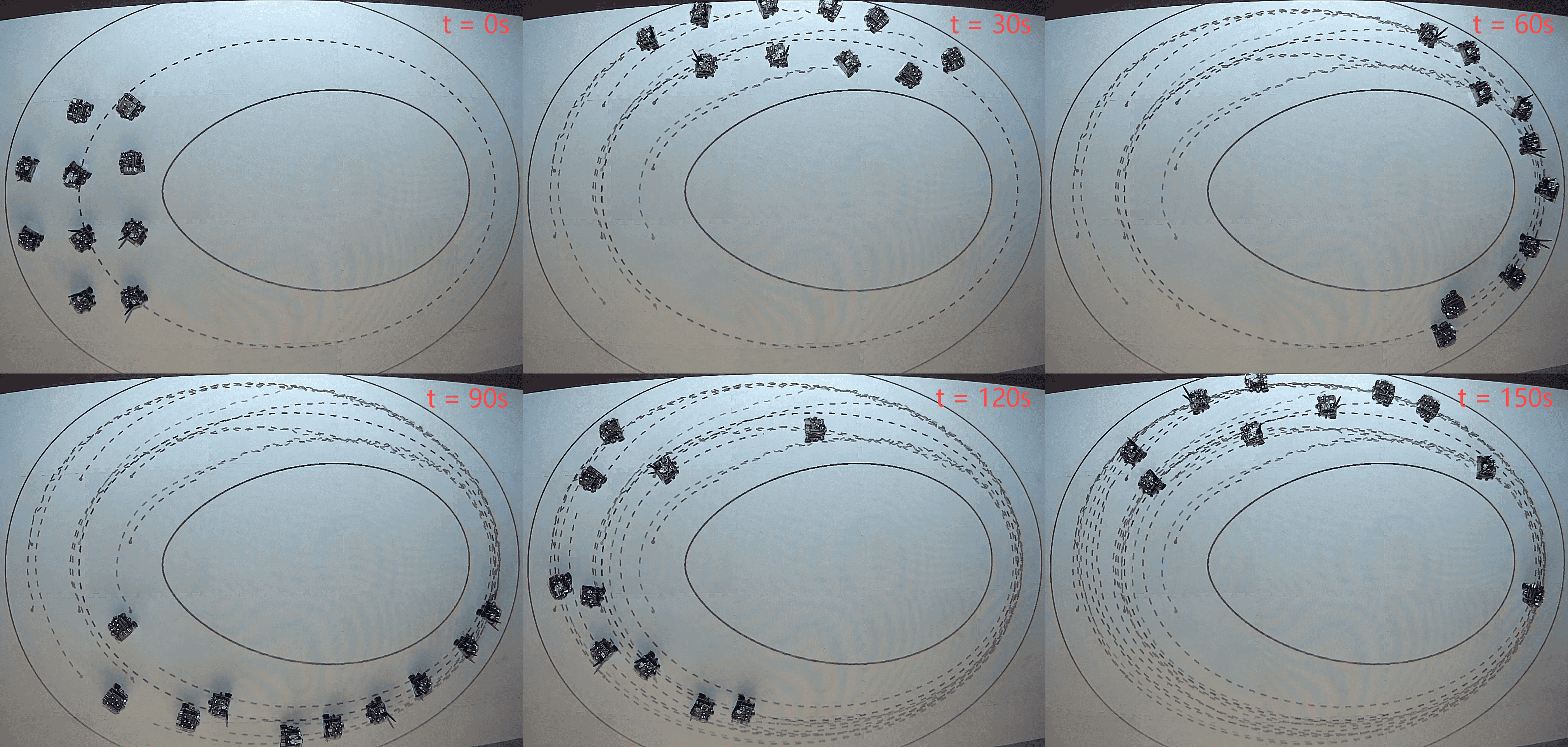}
	\caption{Behavior of the ground mobile robots under the controller $\mathbf{v}_i$}
	\label{F0-150}
\end{figure}

In this subsection, we apply the proposed controller (\ref{controller}) to ground mobile robots, with experiments conducted on the \textit{Robotarium} platform \cite{robotarium1,robotarium2}—a remote multi-robot experimentation platform developed by the Georgia Institute of Technology.
To demonstrate the method’s applicability to diverse virtual tubes, we test swarm navigation in a narrow annular virtual tube, as depicted in Fig. \ref{F0-150}. The safety radius is set to $0.075m$.
% The tube’s cross-section has a minimum radius of is $2r_s = 0.15m$, making it a narrow tube.
% Experimental parameters are provided in Table \ref{parameter_experiment}. 

\begin{figure}[!t] 
	\centering
	\includegraphics[width = 3.5in]{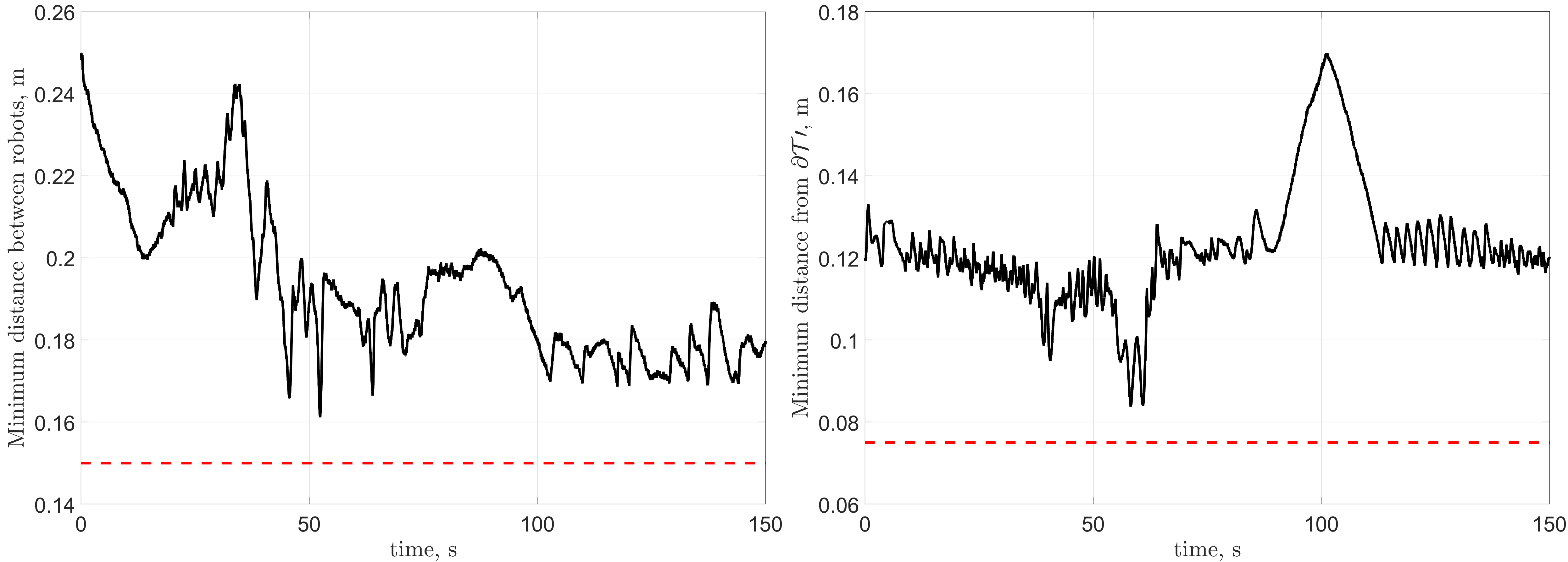}
	\caption{Minimum distance between robots (left) and from $\partial \mathcal{T}'$ (right).}
	\label{robot_tube_distance_experiment}
\end{figure}

As shown in Fig. \ref{F0-150}, $N = 10$ robots start inside the tube with no initial collisions. The experiment lasts 150 seconds, with dashed lines in the figure representing the robots’ motion trajectories. Results confirm that robots avoid collisions and stay within the tube throughout movement. 
Furthermore, Fig. \ref{robot_tube_distance_experiment} shows that inter-robot distances remain greater than $2r_s = 0.15m$, and each robot’s distance to the non-cross-sectional boundary stays greater than $r_s = 0.075m$. This further verifies safe, collision-free navigation in the annular virtual tube.

% \begin{table}[!t]
% 	\centering
% 	\caption{Experiment Parameter Settings}
% 	\begin{tabular}{cccccccc}
% 		\toprule % 上线
% 		$k_1$ & $k_2$ & $k_3$ & $r_s$ & $r_a$ & $h$ & $\varepsilon_m/\varepsilon_s/\varepsilon_t$ & $v_{\max}$ \\
% 		\midrule % 中线
% 		0.05 & 1 & 1 & $0.075m$ & $0.12m$ & 0.15 & $10^{-6}$ & $0.1m/s$ \\
% 		\bottomrule % 下线
% 	\end{tabular}
% 	\label{parameter_experiment}
% \end{table}

% \begin{figure}[!t]
% 	\centering
% 	\subfloat[]{\includegraphics[width=3in]{figures/robot_distance_experiment.png}%
% 			\label{robot_distance_experiment}}
% 	\hfil
% 	\subfloat[]{\includegraphics[width=3in]{figures/tube_distance_experiment.png}%
% 			\label{tube_distance_experiment}}
% 	\caption{(a) Minimum distance between ground mobile robots. (b) Minimum distance from the tube boundary.}
% \end{figure}

\section{Conclusion}
This article investigates spatial configuration control for robot swarms navigating through narrow virtual tubes.
We start by presenting the virtual tube model and its key properties, then design a desired spatial density function to derive a distribution regulation term, and integrate it with safe navigation term to form a saturated velocity command. This ensures collision-free navigation and LISS of density tracking, increasing the swarm’s AMD to enhance safety and efficiency. Simulations and real applications validate the approach’s effectiveness in narrow virtual tube navigation.
Notably, our method uses a hybrid control structure; a key future direction is developing a distributed controller for safe navigation and stable density regulation with only local information.

{\appendix[Proof of Theorem \ref{theorem}]  \label{Appendix}
The proof proceeds in three steps: (1) collision avoidance with robots and $\partial \mathcal{T}'$; (2) all robots reaching $\mathcal{C}(L)$ under (\ref{controller}); (3) LISS of the density tracking error under (\ref{velocity_field3}).

(1) First, by Lemma 2 in \cite{quan2022practical}, all robots avoid inter-robot collisions and collisions with the tube’s non-cross-sectional boundaries during traversal.

(2) Next, consider the Lyapunov-like function
\begin{equation*}\label{Lyapunov0}
	V(t) = \sum_{i=1}^{N} \left( V_{\mathrm{l},i}+\frac{1}{2}\sum_{j=1,j\neq i}^{N}V_{\mathrm{m},ij} + V_{\mathrm{t},i} \right).
\end{equation*} 
Differentiating $V$ then substituting (\ref{controller}) yields
\begin{equation*}\label{dot_Lyapunov0}
	\begin{aligned}
		\dot{V}(t) 
		& = \sum_{i=1}^{N} \bigg( 
		\mathrm{sat} \left( -(L-l_{\mathbf{p}_i}) \eta(\mathbf{p}_i) \mathbf{t}_c(l_{\mathbf{p}_i}), k_1 \right) 
		\\
		& \quad \quad \quad \enspace + \sum_{j\in\mathcal{N}_{\mathrm{m},i}}\frac{\partial V_{\mathrm{m},ij}}{\partial \left\|\tilde{\mathbf{p}}_{\mathrm{m},ij}\right\|}\frac{\tilde{\mathbf{p}}_{\mathrm{m},ij}}{\left\|\tilde{\mathbf{p}}_{\mathrm{m},ij}\right\|}
		+ \frac{\partial V_{\mathrm{t},i}}{\partial \mathbf{p}_i}
		\bigg) \cdot \mathbf{v}_{i}
		\\
		&= \sum_{i=1}^{N} -\kappa_m \mathbf{u}_{123,i} \cdot \left( \mathbf{u}_{123,i} + \mathbf{u}_{4,i} \right)\\
		&\le \sum_{i=1}^{N} -\kappa_m \|\mathbf{u}_{123,i}\|^2 + \kappa_m \|\mathbf{u}_{123,i}\|\|\mathbf{u}_{4,i}\| \overset{(\ref{condition1})}{\le} 0.
	\end{aligned}
\end{equation*}

Following the proof of Theorem 1 in \cite{quan2023distributed}, when condition $\dot{V}(t) \le 0$ holds, the invariant set principle shows there exists $t_1 > 0$ such that $l_{\mathbf{p}_1(t_1)} = L$. 
By Assumption \ref{assumption4}, after the first robot exits the tube, the same analysis applies to remaining robots. Thus, for each robot $\mathbf{p}_i$, $i=1,\cdots,N$, there exists $t_i > 0$ with $l_{\mathbf{p}_i(t_i)} = L$, meaning all robots eventually reach the terminal cross-section $\mathcal{C}(L)$.

(3) Last, we prove the stability of the density tracking error $\Phi$. Consider a candidate LISS Lyapunov function
\begin{equation} \label{Lyapunov_function}
	\tilde{V}(t)=\frac{1}{2} \|\Phi\|_{L^2(\mathcal{T})}^2 = \frac{1}{2} \int_{\mathcal{T}}\Phi^2 \mathrm{d} A.
\end{equation}

According to \cite{lieberman1996second, zheng2021transporting}, there exists the following energy identity $\frac12\int_{\mathcal{T}} \Phi^2 \mathrm{d} A-\frac12\int_\mathcal{T} \Phi_0^2 \mathrm{d} A =\int_0^t\int_{\mathcal{T}}\nabla\Phi \cdot \mathbf{v}(\Phi+\rho_d) \mathrm{d} A \mathrm{d}\tau$.
Then from (\ref{Lyapunov_function}), $\tilde{V}(t)-\tilde{V}(0) = \int_0^t\int_{\mathcal{T}}\nabla\Phi \cdot \mathbf{v}(\Phi+\rho_d) \mathrm{d} A \mathrm{d}\tau$.
Hence, for a.e. $t \in [0, T]$,
\begin{equation} \label{dot_Lyapunov1}
	\dot{\tilde{V}}(t) = \int_{\mathcal{T}}\nabla\Phi \cdot \mathbf{v}(\Phi+\rho_d) \mathrm{d} A = \int_{\mathcal{T}}\nabla\Phi \cdot \mathbf{v}\rho \mathrm{d} A.
\end{equation}

From Theorem 1 in \cite{zheng2021transporting}, $\rho_0 > 0$ implies that $\rho > 0$, hence $\varepsilon$ is well-defined. Rewrite $\rho = \Phi+\rho_d$ and $\hat{\rho} = \rho(1+\varepsilon)$, then substitute (\ref{velocity_field3}) into (\ref{dot_Lyapunov1}). We have
\begin{equation} \label{dot_Lyapunov2}
	\begin{aligned}
		\dot{\tilde{V}}&(t) = -\kappa_m \int_{\mathcal{T}} \nabla\Phi \cdot \left(\rho \frac{\alpha \nabla (\hat{\rho} -\rho_d) -\hat{\rho}\mathbf{w}_e }{\hat{\rho}} \right) \mathrm{d} A \\
		&=\kappa_m \int_{\mathcal{T}} -\nabla\Phi \cdot \frac{\alpha \nabla[\Phi(1+\varepsilon)] + \alpha \nabla(\varepsilon\rho_d) -\hat{\rho} \mathbf{w}_e }{1+\varepsilon} \mathrm{d} A \\
		&=\kappa_m \int_{\mathcal{T}} -\alpha \left|\nabla \Phi\right|^2 \\ 
        & \hspace{40pt} -\nabla \Phi \cdot  \frac{\alpha(\Phi+\rho_d)\nabla\varepsilon + \alpha\varepsilon\nabla\rho_d -\hat{\rho}\mathbf{w}_e }{1+\varepsilon} \mathrm{d} A \\
		&\le\kappa_m \int_{\mathcal{T}} -\alpha \left|\nabla \Phi\right|^2 + \left| \frac{\alpha\Phi\nabla \Phi\cdot \nabla\varepsilon}{1+\varepsilon} \right| + \left| \frac{\alpha\rho_d\nabla \Phi\cdot \nabla\varepsilon}{1+\varepsilon} \right| \\
		&\hspace{40pt} + \left| \frac{\alpha\varepsilon\nabla \Phi\cdot \nabla\rho_d}{1+\varepsilon} \right| + \left| \frac{\nabla \Phi \cdot \hat{\rho}\mathbf{w}_e}{1+\varepsilon}\right| \mathrm{d} A.
	\end{aligned}
\end{equation}
Choose a constant $\theta \in (0,1)$ to decompose $-\alpha \left|\nabla \Phi\right|^2$ into the following form
\begin{equation} \label{decompose}
	\begin{aligned}
		-\alpha \left|\nabla \Phi\right|^2 &= -\alpha(1-\theta)\left|\nabla \Phi\right|^2 - \alpha\theta\left|\nabla \Phi\right|^2\\
		&\le -\alpha_{\min}(1-\theta)\left|\nabla \Phi\right|^2 - \alpha_{\min}\theta\left|\nabla \Phi\right|^2.
	\end{aligned}
\end{equation}

Next, apply the generalized Hölder's inequality to the last four terms of (\ref{dot_Lyapunov2}) and the Poincaré inequality to the first and second terms of (\ref{decompose}). This yields a constant $C>0$ such that
\begin{equation}\label{dot_Lyapunov3}
	\begin{aligned}
		\dot{\tilde{V}}&(t) \le  
		-\frac{\kappa_m\alpha_{\min}(1-\theta)}{C^2} \left\|  \Phi  \right\|_{L^2(\mathcal{T})}^2 \\
		&- \frac{\kappa_m \alpha_{\min}\theta}{C} \left\| \nabla \Phi  \right\|_{L^2(\mathcal{T})}\left\| \Phi  \right\|_{L^2(\mathcal{T})} \\
		& + \kappa_m\|\alpha\|_{L^\infty(\mathcal{T})} \left\| \nabla \Phi  \right\|_{L^2(\mathcal{T})} \left\| \Phi  \right\|_{L^2(\mathcal{T})} \left\| \frac{\nabla \varepsilon}{1+\varepsilon} \right\|_{L^{\infty}(\mathcal{T})} \\
		& + \kappa_m\|\alpha\|_{L^\infty(\mathcal{T})} \left\| \nabla \Phi  \right\|_{L^2(\mathcal{T})} \left\| \rho_d  \right\|_{L^2(\mathcal{T})} \left\| \frac{\nabla \varepsilon}{1+\varepsilon} \right\|_{L^{\infty}(\mathcal{T})}\\
		& + \kappa_m\left\| \nabla \Phi  \right\|_{L^2(\mathcal{T})} \left\| \alpha \nabla \rho_d  \right\|_{L^\infty(\mathcal{T})} \left\| \frac{ \varepsilon}{1+\varepsilon} \right\|_{L^2(\mathcal{T})} \\
		& + \kappa_m\left\| \nabla \Phi  \right\|_{L^2(\mathcal{T})} \left\| \hat{\rho}\mathbf{w}_e \right\|_{L^\infty(\mathcal{T})} \left\| \frac{1}{1+\varepsilon} \right\|_{L^2(\mathcal{T})}.
	\end{aligned}
\end{equation}

Note that collision avoidance has been established in step (1), preventing singularities in $\mathbf{u}_{2,i}$ and $\mathbf{u}_{3,i}$, and keeping them bounded. With $\mathbf{u}_{1,i}$ being saturated, $\mathbf{u}_{123,i}$ is bounded. Thus $\mathbf{w}_e$ in (\ref{velocity_field3}) can be chosen to ensure $\left\| \hat{\rho}\mathbf{w}_e \right\|_{L^\infty(\mathcal{T})} < \infty$.
Now we can assert: If $\Phi$ satisfies  
\begin{equation} \label{condition3}
	\begin{aligned}
		&\left\| \Phi \right\|_{L^2(\mathcal{T})} \\
		& \hspace{10pt} \ge \frac{ \|\alpha\|_{L^\infty(\mathcal{T})} \left\| \rho_d  \right\|_{L^2(\mathcal{T})}d(t) + \left\| \alpha\nabla \rho_d  \right\|_{L^2(\mathcal{T})}d(t) }
		{ \frac{\alpha_{\min} \theta}{C} - \|\alpha\|_{L^\infty(\mathcal{T}) } d(t) } \\
		&\hspace{10pt} + \frac{\left\| \hat{\rho}\mathbf{w}_e \right\|_{L^\infty (\mathcal{T})}d(t) }{\frac{\alpha_{\min} \theta}{C} - \|\alpha\|_{L^\infty(\mathcal{T}) } d(t)} =: \chi \left( d(t)\right),
	\end{aligned}
\end{equation}
then 
\begin{equation} \label{assertion1}
	\dot{\tilde{V}}(t) \le -\frac{\kappa_m\alpha_{\min} (1-\theta)}{C^2} \left\|  \Phi  \right\|_{L^2(\mathcal{T})}^2 =: -W \left( \left\| \Phi  \right\|_{L^2(\mathcal{T})}^2 \right),
\end{equation}
where $d(t) = \max \{ \|\frac{\nabla\varepsilon}{1+\varepsilon}\|_{L^\infty(\mathcal{T})}, \|\frac\varepsilon{1+\varepsilon}\|_{L^2(\mathcal{T})}, \|\frac1{1+\varepsilon}\|_{L^2(\mathcal{T})} \}.$
% \begin{equation*}
% 	\begin{aligned}
% 		d(t) = \max \{ &\|\frac{\nabla\varepsilon}{1+\varepsilon}\|_{L^\infty(\mathcal{T})}(t), \|\frac\varepsilon{1+\varepsilon}\|_{L^2(\mathcal{T})}(t), \\
% 		&\|\frac1{1+\varepsilon}\|_{L^2(\mathcal{T})}(t) \}.
% 	\end{aligned}
% \end{equation*}

To be specific, if (\ref{condition2}) holds, with (\ref{condition3}), then 
\begin{equation*} \label{assertion2}
	\begin{aligned}
		&\left\| \Phi \right\|_{L^2(\mathcal{T})} \ge \frac{ \|\alpha\|_{L^\infty(\mathcal{T})} \left\| \rho_d  \right\|_{L^2(\mathcal{T})} \|\frac{\nabla\varepsilon}{1+\varepsilon}\|_{L^\infty(\mathcal{T})} }{\frac{\alpha_{\min} \theta}{C}-\|\alpha\|_{L^\infty(\mathcal{T})}\|\frac{\nabla\varepsilon}{1+\varepsilon}\|_{L^\infty(\mathcal{T})} } \\
		&\hspace{10pt} + \frac{ \left\| \alpha \nabla \rho_d  \right\|_{L^\infty(\mathcal{T})} \|\frac{\varepsilon}{1+\varepsilon}\|_{L^2(\mathcal{T})} + \left\| \hat{\rho}\mathbf{w}_e \right\|_{L^\infty(\mathcal{T})}\|\frac{1}{1+\varepsilon}\|_{L^2(\mathcal{T})} }{\frac{\alpha_{\min} \theta}{C}-\|\alpha\|_{L^\infty(\mathcal{T})}\|\frac{\nabla\varepsilon}{1+\varepsilon}\|_{L^\infty(\mathcal{T})} }.
	\end{aligned}
\end{equation*}
This obviously derives the following inequality:
$\frac{\alpha_{\min}\theta}{C} \left\| \Phi  \right\|_{L^2(\mathcal{T})} \ge \|\alpha\|_{L^\infty(\mathcal{T})} \left\| \rho_d  \right\|_{L^2(\mathcal{T})} \left\| \frac{\nabla \varepsilon}{1+\varepsilon} \right\|_{L^{\infty}(\mathcal{T})}  
+ \left\|\alpha \nabla \rho_d  \right\|_{L^{\infty}(\mathcal{T})} \left\| \frac{ \varepsilon}{1+\varepsilon} \right\|_{L^2(\mathcal{T})}  
+ \left\| \hat{\rho}\mathbf{w}_e\right\|_{L^{\infty}(\mathcal{T})} \left\| \frac{1}{1+\varepsilon} \right\|_{L^2(\mathcal{T})}
+ \|\alpha\|_{L^\infty(\mathcal{T})} \left\| \Phi  \right\|_{L^2(\mathcal{T})} \left\| \frac{\nabla \varepsilon}{1+\varepsilon} \right\|_{L^{\infty}(\mathcal{T})}$.
% \begin{equation} \label{assertion3}
% 	\begin{aligned}
% 		\frac{\alpha_{\min}\theta}{C} \left\| \Phi  \right\|_{L^2(\mathcal{T})} 
% 		&\ge  \|\alpha\|_{L^\infty(\mathcal{T})} \left\| \Phi  \right\|_{L^2(\mathcal{T})} \left\| \frac{\nabla \varepsilon}{1+\varepsilon} \right\|_{L^{\infty}(\mathcal{T})} \\
% 		&\hspace{10pt} + \|\alpha\|_{L^\infty(\mathcal{T})} \left\| \rho_d  \right\|_{L^2(\mathcal{T})} \left\| \frac{\nabla \varepsilon}{1+\varepsilon} \right\|_{L^{\infty}(\mathcal{T})} \\
% 		&\hspace{10pt} + \left\|\alpha \nabla \rho_d  \right\|_{L^{\infty}(\mathcal{T})} \left\| \frac{ \varepsilon}{1+\varepsilon} \right\|_{L^2(\mathcal{T})} \\
% 		&\hspace{10pt} + \left\| \hat{\rho}\mathbf{w}_e\right\|_{L^{\infty}(\mathcal{T})} \left\| \frac{1}{1+\varepsilon} \right\|_{L^2(\mathcal{T})}.
% 	\end{aligned}
% \end{equation}

With this inequality, (\ref{assertion1}) can be obtained from (\ref{dot_Lyapunov3}), thus the assertion holds. Furthermore, due to the fact that $W$ in (\ref{assertion1}) is positive definite and $\chi$ in (\ref{condition3}) belongs to $\mathcal{K} := \{ f: \mathbb{R}_+ \to \mathbb{R}_+  \,\, | \,\, f \,\, \mathrm{is\,\, continuous\,\, and\,\, strictly} \,\,\mathrm{increasing\,\,  with\,\, \textit{f}(0)=0} \}$,
$\tilde{V}$ is indeed an LISS Lyapunov function for (\ref{error_system}). Thus by Theorem 3 in \cite{zheng2021transporting}, $\Phi$ is LISS with respect to $d(t)$. 	
}

\end{document}